\documentclass[a4paper,12pt]{article}

\usepackage{comment}
\usepackage{fancyvrb}
\usepackage{graphicx}
\usepackage{hyperref}
\usepackage[ragged]{footmisc}
\usepackage{ragged2e}
\usepackage[section]{placeins}
\usepackage{gensymb}

\begin{document}

\title{The SP theory of intelligence and the representation and processing of knowledge in the brain}

\author{J Gerard Wolff\footnote{Dr Gerry Wolff, BA (Cantab), PhD (Wales), CEng, MBCS (CITP); CognitionResearch.org, Menai Bridge, UK; \href{mailto:jgw@cognitionresearch.org}{jgw@cognitionresearch.org}; +44 (0) 1248 712962; +44 (0) 7746 290775; {\em Skype}: gerry.wolff; {\em Web}: \href{http://www.cognitionresearch.org}{www.cognitionresearch.org}.}}

\maketitle

\begin{abstract}

The {\em SP theory of intelligence}, with its realisation in the {\em SP computer model}, aims to simplify and integrate observations and concepts across artificial intelligence, mainstream computing, mathematics, and human perception and cognition, with information compression as a unifying theme. This paper describes how abstract structures and processes in the theory may be realised in terms of neurons, their interconnections, and the transmission of signals between neurons. This part of the SP theory---{\em SP-neural}---is a tentative and partial model for the representation and processing of knowledge in the brain. Empirical support for the SP theory---outlined in the paper---provides indirect support for SP-neural.

In the SP theory (apart from SP-neural), all kinds of knowledge are represented with {\em patterns}, where a pattern is an array of atomic symbols in one or two dimensions. In SP-neural, the concept of a `pattern' is realised as an array of neurons called a {\em pattern assembly}, similar to Hebb's concept of a `cell assembly' but with important differences.

Central to the processing of information in the SP system is the powerful concept of {\em multiple alignment}, borrowed and adapted from bioinformatics. Processes such as pattern recognition, reasoning and problem solving are achieved via the building of multiple alignments, while unsupervised learning is achieved by creating patterns from sensory information and also by creating patterns from multiple alignments in which there is a partial match between one pattern and another.

It is envisaged that, in SP-neural, short-lived neural structures equivalent to multiple alignments will be created via an inter-play of excitatory and inhibitory neural signals. It is also envisaged that unsupervised learning will be achieved by the creation of pattern assemblies from sensory information and from the neural equivalents of multiple alignments, much as in the non-neural SP theory---and significantly different from the `Hebbian' kinds of learning which are widely used in the kinds of artificial neural network that are popular in computer science.

The paper discusses several associated issues, with relevant empirical evidence.

\end{abstract}


\section{Introduction}


The {\em SP theory of intelligence}, and its realisation in the {\em SP computer model}, is a unique attempt to simplify and integrate observations and concepts across artificial intelligence, mainstream computing, mathematics, and human perception and cognition, with information compression as a unifying theme.

This paper, which derives from \cite[chapter 11]{wolff_2006} with revisions and updates, describes how abstract structures and processes in the SP theory may be realised in terms of neurons, their interconnections, and the transmission of impulses between neurons. This part of the SP theory---called {\em SP-neural}---may be seen as a tentative and partial theory of the representation and processing of knowledge in the brain. As such, it may prove useful as a source of ideas for theoretical and empirical investigations in the future. For the sake of clarity, the abstract parts of the theory, excluding SP-neural, will be referred to as ``SP-abstract''.

It is envisaged that SP-neural will be further developed in the form of a computer model. As with the existing computer model of SP-abstract, the development of this new computer model will help to guard against vagueness in the theory, it will serve as a means of testing ideas to see whether or not they work as anticipated, and it will be a means of demonstrating what the model can do, and validating it against empirical data.

The next section says something about the theoretical orientation of this research. Then SP-abstract will be described briefly as a foundation for the several sections that follow which describe aspects of SP-neural and associated issues.

\section{Theoretical orientation}\label{theoretical_orientation_section}

Cosmologist John Barrow has written that ``Science is, at root, just the search for compression in the world'' \cite[p.~247]{barrow_1992}, an idea which may be seen to be equivalent to Occam's Razor---a good theory should combine conceptual {\em simplicity} with descriptive or explanatory {\em power}. This is because compression of any given body of information, {\bf I}, may be seen as a process of reducing `redundancy' of information in {\bf I} and thus increasing its `simplicity', whilst retaining as much as possible of its non-redundant descriptive and explanatory `power'.

This works best when {\bf I} is large. But this has not always been observed in practice: Newell \cite[p.~303]{newell_1973} urged researchers in psychology to address ``a genuine slab of human behaviour''; and McCorduck \cite[pp.~417 and 424]{mccorduck_2004} has described how research in artificial intelligence became fragmented into many narrow sub-fields.

In the light of these observations, and in the spirit of research on ``unified theories of cognition'' \cite{newell_1990} and ``artificial general intelligence'',\footnote{See, for example, ``Artificial General Intelligence'', {\em Wikipedia}, \href{http://bit.ly/1ZxCQPo}{bit.ly/1ZxCQPo}, retrieved 2016-01-19.} the SP programme of research has attempted to simplify and integrate observations and concepts across a broad canvass, resisting the temptation to concentrate only on one small area.

In connection with these ideas, the name ``SP'' may be seen to be short for {\em simplicity} and {\em power}. This is partly because, as we shall see, the SP theory, despite its relative simplicity, has quite a lot to say about a wide range of observations and concepts. But more importantly it is because information compression lies at the heart of how the SP system works.

\section{SP-abstract in brief}\label{sp-abstract_outline_section}

As a basis for the description of SP-neural, this section provides a brief informal account of SP-abstract. The theory is described most fully in \cite{wolff_2006} and quite fully but more briefly in \cite{sp_extended_overview}. Details of other publications in the SP programme, many of them with download links, are shown on the website of CognitionResearch.org (\href{http://bit.ly/1mSs5XT}{bit.ly/1mSs5XT}).

\subsection{Origins and foundations of the SP theory}

The origins of SP theory are mainly in a body of research by Attneave \cite{attneave_1954}, Barlow \cite{barlow_1959,barlow_1969} and others suggesting that much of the workings of brains and nervous systems may be understood as compression of information, and my own research (summarised in \cite{wolff_1988}) suggesting that, to a large extent, language learning may be understood in the same terms. There is more about the foundations of the theory in \cite{sp_foundations}.

\subsection{Elements of SP-abstract}\label{elements_of_sp-a_section}

In SP-abstract, all kinds of knowledge are represented with {\em patterns}, where a pattern is an array of atomic {\em symbols} in one or two dimensions. At present, the SP computer model\footnote{The current version of the SP computer model is SP71, the source code for which may be downloaded via a link from \href{http://bit.ly/1ZHUqPE}{bit.ly/1ZHUqPE}. This version of the computer model is very similar to SP70, described in \cite[Sections 3.9.2 and 9.2]{wolff_2006}.} works only with 1D patterns but it is envisaged that the model will be generalised to work with 2D patterns. In this connection, a `symbol' is simply a `mark' that can make a yes/no match with any other symbol---no other result is permitted.

In most of the examples shown in this paper, symbols are shown as alphanumeric characters or short strings of characters but, when the SP system is used to model biological structures and processes, such representations may be interpreted as low-level elements of perception such as formants or formant ratios in the case of speech or lines and junctions between lines in the case of vision (see also Section \ref{sp-n_sensory_data_receptor_array_section}).

To help cut through mathematical complexities associated with information compression, the SP system---SP-abstract and its realisation in the SP computer model---is founded on a simple, `primitive' idea: that information may be compressed by finding full or partial matches between patterns and merging or `unifying' the parts that are the same. This principle---``Information Compression via the Matching and Unification of Patterns'' (ICMUP)---provides the foundation for a powerful concept of {\em multiple alignment}, borrowed and adapted from bioinformatics. The multiple alignment concept, outlined in Section \ref{sp-a_multiple_alignment_section}, below, is itself central in the workings of SP-abstract and is the key to versatility and adaptability of the SP system. It has the potential to be as significant for the understanding of `intelligence' in a broad sense as is DNA for biological sciences.

\subsection{SP patterns, multiple alignment, and the representation and processing of knowledge}

In themselves, SP patterns are not very expressive. But in the multiple alignment framework (Section \ref{sp-a_multiple_alignment_section}) they become a very versatile medium for the representation of diverse forms of knowledge. And the building of multiple alignments, together with processes for unsupervised learning (Sections \ref{sp-a_early_learning_section} and \ref{sp-a_later_learning_section}), has proved to be a powerful means of modelling diverse kinds of processing.

The two things together---SP patterns and multiple alignment---have the potential to be a ``universal framework for the representation and processing of diverse kinds of knowledge'' (UFK), as discussed in \cite{sp_big_data}.

An implication of these ideas is that there would not, for example, be any difference between the representation and processing of non-syntactic cognitive knowledge and the representation and processing of the syntactic forms of natural language. A framework that can accommodate both kinds of knowledge is likely to facilitate their seamless integration, as discussed in Section \ref{sp-a_evaluation_systems_section}.

\subsection{Early stages of learning}\label{sp-a_early_learning_section}

The SP theory is conceived as a brain-like system that receives {\em New} patterns via its `senses' and stores some or all of them, in compressed form, as {\em Old} patterns. In broad terms, this is how the system learns.

In the SP system, all learning is `unsupervised',\footnote{See ``Unsupervised learning'', {\em Wikipedia}, \href{http://bit.ly/22nEPL2}{bit.ly/22nEPL2}, retrieved 2016-03-17} meaning that it does not depend on assistance by a `teacher', the grading of learning materials from simple to complex, or the provision of `negative' examples of concepts to be learned---meaning examples that are marked as `wrong' ({\em cf}.~\cite{gold_1967}). Notwithstanding the importance of schools and colleges, it appears that most human learning is unsupervised. Other kinds of learning, such as `supervised' learning (learning from labelled examples),\footnote{See ``Supervised learning'', {\em Wikipedia}, \href{http://bit.ly/1nR4ybK}{bit.ly/1nR4ybK}, retrieved 2016-03-17.} or `reinforcement' learning (learning with carrots and sticks),\footnote{See ``Reinforcement learning'', {\em Wikipedia}, \href{http://bit.ly/1R0RoDv}{bit.ly/1R0RoDv}, retrieved 2016-03-17.} may be seen as special cases of unsupervised learning \cite[Section V]{sp_autonomous_robots}.

At the beginning of processing by the system, when the repository of Old patterns is empty,\footnote{Although it is likely that, contrary to what Noam Chomsky and others have suggested, a newborn child does {\em not} have any kind of detailed knowledge of the structure of natural language, it {\em is} likely he or she does have inborn knowledge such as how to suck milk from a breast. In this respect (and others), the SP theory, insofar it is seen as a model of human cognition, is not entirely accurate.} New patterns are stored as they are received but with the addition of system-generated `ID' symbols at the beginning and end. For example, a New pattern like `\texttt{t h e b i g h o u s e}' would be stored as an Old pattern like `\texttt{A 1 t h e b i g h o u s e \#A}'. Here, the lower-case letters are atomic symbols that may represent actual letters but could represent basic elements of speech (such as formant ratios or formant transitions), or basic elements of vision (such as lines or corners), and likewise with other sensory data.

Later, when some Old patterns have been stored, the system may start to recognise full or partial matches between New and Old patterns. If a New pattern is exactly the same as an Old pattern (excluding the ID-symbols), then frequency measures for that pattern and its constituent symbols are incremented. These measures, which are continually updated at all stages of processing, have an important role to play in calculating probabilities of structures and inferences and in guiding the processes of building multiple alignments (Section \ref{sp-a_multiple_alignment_section}) and unsupervised learning.

With partial matches, the system will form multiple alignments like the one shown in Figure \ref{partial_match_figure}, with a New pattern in row 0 and an Old pattern in row 1.

\begin{figure}[!htbp]
\fontsize{10.00pt}{12.00pt}
\centering
{\bf
\begin{BVerbatim}
0     t h e       s m a l l h o u s e    0
      | | |                 | | | | |
1 A 1 t h e b i g           h o u s e #A 1
\end{BVerbatim}
}
\caption{A multiple alignment produced by the SP computer model showing a partial match between a New pattern (in row 0) and an Old pattern (in row 1).}
\label{partial_match_figure}
\end{figure}

From a partial match like this, the system creates Old patterns from the parts that match each other and from the parts that don't. Each newly-created Old pattern will be given system-generated ID-symbols. The result in this case would be patterns like these: `\texttt{B 1 t h e \#B}', `\texttt{C 1 h o u s e \#C}', `\texttt{D 1 s m a l l \#D}', `\texttt{D 2 b i g \#D}'. In addition, the system forms an abstract pattern like this: `\texttt{E 1 B \#B D \#D C \#C \#E}' which records the sequence [`\texttt{B 1 t h e \#B}', (`\texttt{D 1 s m a l l \#D}' or `\texttt{D 2 b i g \#D}'), `\texttt{C 1 h o u s e \#C}'] in terms the ID-symbols of the constituent patterns.

Notice how `\texttt{s m a l l}' and `\texttt{b i g}' have both been given the ID-symbol `\texttt{D}' at their beginnings and the ID-symbol `\texttt{\#D}' at their ends. These additions, coupled with the use of the same two ID-symbols in the abstract pattern `\texttt{E 1 B \#B D \#D C \#C \#E}' has the effect of assigning `\texttt{s m a l l}' and `\texttt{b i g}' to the same syntactic category, which looks like the beginnings of the `adjective' part of speech.

The overall result in this example is a collection of SP patterns that functions as a simple grammar to describe the phrases {\em the small house} and {\em the big house}.

In practice, the SP computer model may form many other multiple alignments, patterns and grammars which are much less tidy than the ones shown. But, as outlined in Sections \ref{sp-a_multiple_alignment_section} and \ref{sp-a_later_learning_section}, the system is able to home in on structures that are `good' in terms of information compression.

As we shall see (Sections \ref{sp-a_multiple_alignment_section}, \ref{sp-a_evaluation_theory_section}, and \ref{sp-n_non-syntactic_knowledge_section}), SP patterns, within the SP system, are remarkably versatile and expressive, with at least the power of context-sensitive grammars \cite[Chapter 5]{wolff_2006}.

\subsection{The multiple alignment concept}\label{sp-a_multiple_alignment_section}

\begin{figure}[!htbp]
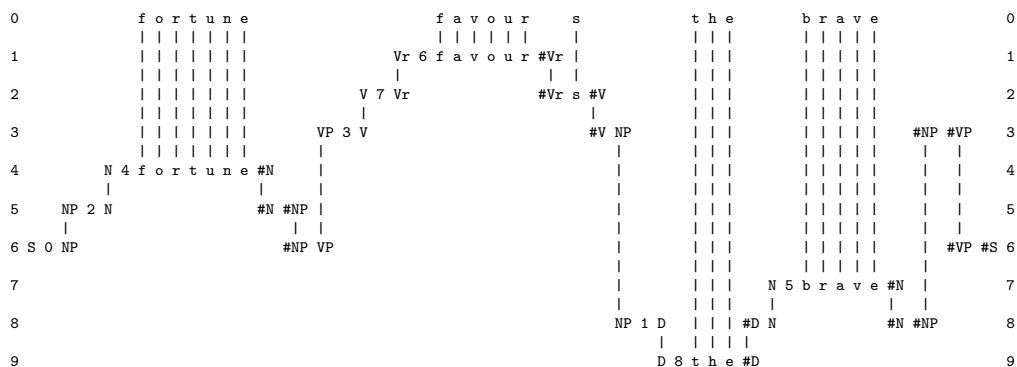

\fontsize{06.00pt}{07.20pt}
\centering
{\bf
\begin{BVerbatim}
0              f o r t u n e                      f a v o u r     s             t h e        b r a v e               0
               | | | | | | |                      | | | | | |     |             | | |        | | | | |
1              | | | | | | |                 Vr 6 f a v o u r #Vr |             | | |        | | | | |               1
               | | | | | | |                 |                 |  |             | | |        | | | | |
2              | | | | | | |             V 7 Vr               #Vr s #V          | | |        | | | | |               2
               | | | | | | |             |                          |           | | |        | | | | |
3              | | | | | | |        VP 3 V                          #V NP       | | |        | | | | |    #NP #VP    3
               | | | | | | |        |                                  |        | | |        | | | | |     |   |
4          N 4 f o r t u n e #N     |                                  |        | | |        | | | | |     |   |     4
           |                 |      |                                  |        | | |        | | | | |     |   |
5     NP 2 N                 #N #NP |                                  |        | | |        | | | | |     |   |     5
      |                          |  |                                  |        | | |        | | | | |     |   |
6 S 0 NP                        #NP VP                                 |        | | |        | | | | |     |  #VP #S 6
                                                                       |        | | |        | | | | |     |
7                                                                      |        | | |    N 5 b r a v e #N  |         7
                                                                       |        | | |    |             |   |
8                                                                      NP 1 D   | | | #D N             #N #NP        8
                                                                            |   | | | |
9                                                                           D 8 t h e #D                             9
\end{BVerbatim}
}
\caption{The best multiple alignment produced by the SP computer model with a New pattern representing a sentence to be parsed and a repository of user-supplied Old patterns representing grammatical categories, including words. In the multiple alignment, the New pattern appears in row 0 and some of the Old patterns supplied to the system appear in rows 1 to 9, one pattern per row.}
\label{fortune_brave_multiple_alignment_figure}
\end{figure}

The multiple alignment shown in Figure \ref{partial_match_figure} is unusually simple because it contains only two patterns. More commonly, the system forms `good' multiple alignments like the one shown in Figure \ref{fortune_brave_multiple_alignment_figure}, with one New pattern (in row 0) and several Old patterns (one in each of several other rows).\footnote{In this case, the SP computer model was supplied with an appropriate set of Old patterns. It did not learn them for itself.} As a matter of convention, the New pattern is always shown in row 0, but the order of the Old patterns across the other rows is not significant.

A multiple alignment like the one shown in Figure \ref{fortune_brave_multiple_alignment_figure} is built in stages, using heuristic search at each stage to weed out structures that are `bad' in terms of information compression and retaining those that are `good'. Problems of computational complexity are reduced or eliminated by a scaling back of ambition: instead of searching for theoretically-ideal solutions, one merely searches for solutions that are ``good enough''.

In this example, multiple alignment achieves the effect of parsing the sentence into parts and sub-parts, such as a sentence (`S') defined by the pattern in row 6, one kind of noun phrase (`NP') defined by the pattern that appears in row 5, and another kind of noun phrase shown in row 8, a verb phrase (`VP') defined by the pattern in row 3, nouns (`N') defined by the patterns in rows 4 and 7, and so on. But there is much more than this to the multiple alignment concept as it has been developed in the SP programme. It turns out to be a remarkably versatile framework for the representation and processing of diverse kinds of knowledge---non-verbal patterns and pattern recognition, logical and probabilistic kinds of `rules' and several kinds of reasoning, and more (Sections \ref{sp-a_evaluation_theory_section} and \ref{sp-n_non-syntactic_knowledge_section}).

A point worth mentioning here is that, although the multiple concept is entirely non-hierarchical, it can model several kinds of hierarchy and heterarchy (Section \ref{sp-a_evaluation_theory_section}), as illustrated by the parsing example in Figure \ref{fortune_brave_multiple_alignment_figure}. And such hierarchies may not always be `strict' hierarchies because any pattern may be aligned with any other pattern and, within one multiple alignment, any pattern may be aligned with two or more other patterns.

\subsection{Deriving a code pattern from a multiple alignment}\label{sp-a_deriving_code_pattern_section}

From a multiple alignment like the one shown in Figure \ref{fortune_brave_multiple_alignment_figure}, the SP system may derive a {\em code pattern}---a compressed encoding of the sentence---as follows: scan the multiple alignment from left to right, identifying the ID-symbols that are {\em not} matched with any other symbol and create an SP pattern from the sequence of such symbols. In this case, the result is the pattern `\texttt{S 0 2 4 3 7 6 1 5 \#S}'. This code pattern has several existing or potential uses including:

\begin{itemize}

  \item It provides a basis for calculating a `compression score' for the Old patterns in the multiple alignment, meaning their effectiveness as a means of compressing the New pattern. Compression scores like that have a role in sifting out one or more `good' grammars for any given set of New patterns.

  \item If the code pattern is treated as a New pattern then, with the same Old patterns as when the code pattern was produced, the SP system can recreate the original sentence, as described in Section \ref{sp-n_output_section}.

  \item When SP-abstract is developed to take account of meanings as well as syntax, it is likely that each ID-symbol in the code pattern will take on a dual role: representing each syntactic form (word or other grammatical structure) and representing the meaning of the given syntactic form.

  \item It is envisaged that, with further development of the SP computer model, code patterns will enter into the learning process, as outlined in Section \ref{sp-a_later_learning_section}, next.

\end{itemize}

\subsection{Later stages of learning}\label{sp-a_later_learning_section}

As we saw in Section \ref{sp-a_early_learning_section}, the earliest stage of learning in SP-neural---when the repository of Old patterns is empty or nearly so---is largely a matter of absorbing New information directly with little modification except for the addition of system-generated ID-symbols. Later, when there are more Old patterns in store, the system begins to create Old patterns from partial matches between New and Old patterns. Part of this process is the creation of abstract patterns that describe sequences of lower-level patterns.

As the system begins to create abstract patterns, it will also begin to form multiple alignments like the one shown in Figure \ref{fortune_brave_multiple_alignment_figure}. And, as it begins to form multiple alignments like that, it will also begin to form code patterns, as described in Section \ref{sp-a_deriving_code_pattern_section}.

At all stages of learning, but most prominent in the later stages, is a process of inferring one or more {\em grammars} that are `good' in terms of their ability to encode economically all the New patterns that have been presented to the system. Here, a `grammar' is simply a collection of SP patterns. The term `grammar' has been adopted partly because of the origins of the SP system in research on the learning of natural language \cite{wolff_1988} and partly because the term has come to be used in areas outside computational linguistics, such as pattern recognition.

Inferring grammars that are good in terms of information compression is, like the building multiple alignments, a stage-by-stage process of heuristic search through the vast abstract space of alternatives, discarding `bad' alternatives at each stage, and retaining a few that are `good'. As with the building of multiple alignments, the search aims to find solutions that are ``good enough'', and not necessarily perfect.

It is envisaged that the SP computer model will be developed so that, in this later phase of learning, learning processes will be applied to code patterns as well as to New patterns. It is anticipated that this may overcome two weaknesses in the SP computer model as it is now: that, while it forms abstract patterns at the highest level, it does not form abstract patterns at intermediate levels; and that it does not recognise discontinuous dependencies in knowledge \cite[Section 3.3]{sp_extended_overview}.

In \cite[Chapter 9]{wolff_2006}, there is a much fuller account of unsupervised learning in the SP computer model.

\subsection{Evaluation of SP-abstract}

The SP theory in its abstract form may be evaluated in terms of `simplicity' and `power' of the theory itself (discussed in Section \ref{sp-a_evaluation_theory_section} next), in terms its potential to promote simplification and integration of structures and functions in natural or artificial systems that conform to the theory (Section \ref{sp-a_evaluation_systems_section} below), and in comparison with other AI-related systems.

\subsubsection{Simplicity and power}\label{sp-a_evaluation_theory_section}

In terms of the principles outlined in Section \ref{theoretical_orientation_section}, the SP system, with multiple alignment centre stage, scores well. One relatively simple framework has strengths and potential in the representation of several different kinds of knowledge, in several different aspects of AI, and it has several potential benefits and applications:

\begin{itemize}

    \item {\em Representation and processing of diverse kinds of knowledge}. The SP system (SP-abstract) has strengths and potential in the representation and processing of: class hierarchies and heterarchies, part-whole hierarchies and heterarchies, networks and trees, relational knowledge, rules used in several kinds of reasoning, patterns with pattern recognition, images with the processing of images \cite{sp_vision}, structures in planning and problem solving, structures in three dimensions \cite[Section 6]{sp_vision}, knowledge of sequential and parallel procedures \cite[Section IV-H]{sp_autonomous_robots}. It may also provide an interpretive framework for structures and processes in mathematics \cite[Section 10]{sp_foundations}.

        There is a fuller summary in \cite[Section III-B]{sp_big_data} and much more detail in \cite{wolff_2006,sp_extended_overview}.

    \item {\em Strengths and potential in AI}. The SP theory has things to about several different aspects of AI, as described most fully in \cite{wolff_2006} and more briefly in \cite{sp_extended_overview}. In addition to its capabilities in parsing, described above, the SP system has strengths and potential in the production of natural language, the representation and processing of diverse kinds of semantic structures, the integration of syntax and semantics, pattern recognition, computer vision and modelling aspects of natural vision \cite{sp_vision}, information retrieval, planning, problem solving, and several kinds of reasoning (one-step `deductive' reasoning; abductive reasoning; reasoning with probabilistic decision networks and decision trees; reasoning with `rules'; nonmonotonic reasoning and reasoning with default values; reasoning in Bayesian networks, including `explaining away'; causal diagnosis; reasoning which is not supported by evidence; and inheritance of attributes in an object-oriented class hierarchy or heterarchy). There is also potential for spatial reasoning \cite[Section IV-F.1]{sp_autonomous_robots} and what-if reasoning \cite[Section IV-F.2]{sp_autonomous_robots}. The system also has strengths and potential in unsupervised learning \cite[Chapter 9]{wolff_2006}.

    \item {\em Many potential benefits and applications}. Potential benefits and applications of the SP system include: helping to solve nine problems associated with big data \cite{sp_big_data}; the development of intelligence in autonomous robots, with potential for gains in computational efficiency \cite{sp_autonomous_robots}; the development of computer vision \cite{sp_vision}; it may serve as a versatile database management system, with intelligence \cite{wolff_sp_intelligent_database}; it may serve as an aid in medical diagnosis \cite{wolff_medical_diagnosis}; and there are several other potential benefits and applications, some of which are described in \cite{sp_benefits_apps}.

\end{itemize}

In short, the SP theory, in accordance with Occam's Razor, demonstrates a favourable combination of simplicity and power across a broad canvass. As in other areas of science, this should increase our confidence in the generality of the theory.

\subsubsection{Simplification and integration}\label{sp-a_evaluation_systems_section}

Closely related to simplicity and power in the SP theory are two potential benefits arising from the use of one simple format (SP patterns) for all kinds of knowledge and one relatively simple framework (chiefly multiple alignment) for the processing of all kinds of knowledge:

\begin{itemize}

    \item {\em Simplification}. Those two features (one simple format for knowledge and one simple framework for processing it) can mean substantial simplification of natural systems (brains) and artificial systems (computers) for processing information. The general idea is that one relatively simple system can serve many different functions. In natural systems, there is a potential advantage in terms of natural selection, and in artificial systems there are potential advantages in terms of costs.

    \item {\em Integration}. The same two features are likely to facilitate the seamless integration of diverse kinds of knowledge and diverse aspects of intelligence---pattern recognition, several kinds of reasoning, unsupervised learning, and so on---in any combination, in both natural and artificial systems. It appears that that kind of seamless integration is a key part of the versatility and adaptability of human intelligence and that it will be essential if we are to achieve human-like versatility and adaptability of intelligence in artificial systems.

\end{itemize}

With regard to the seamless integration of diverse kinds of knowledge, this is clearly needed in the understanding and production of natural language. To understand what someone is saying or writing, we obviously need to be able to connect words and syntactic structures with their non-syntactic meanings, and likewise, in reverse, when we write or speak to convey some meaning.

This has not yet been explored in any depth with the SP-abstract conceptual framework but preliminary trials with the SP computer model suggest that it is indeed possible to define syntactic-semantic structures in a set of SP patterns and then, with those patterns playing the role of Old patterns, to analyse a sample sentence and to derive its meanings \cite[Section 5.7, Figure 5.18]{wolff_2006}, and, in a separate exercise with the same set of Old patterns, to derive the same sentence from a representation of its meanings ({\em ibid.}, Figure 5.19).

\subsubsection{Distinctive features and advantages of the SP system compared with other AI-related systems}\label{sp-a_distinctive_features_advantages_section}

In several publications, such as \cite{sp_benefits_apps,wolff_medical_diagnosis,wolff_sp_intelligent_database}, potential benefits and applications of the SP system have been described.

More recently, it has seemed appropriate to say what distinguishes the SP system from other AI-related systems and, more importantly, to describe advantages of the SP system compared AI-related alternatives. Those points have now been set out in some detail in {\em The SP theory of intelligence: its distinctive features and advantages} \cite{sp_alternatives}. It is pertinent to mention that Section V of that paper discusses, in some detail, problems with `deep learning in neural networks' and shows how, in the SP system, they are overcome.

Since many AI-related systems may also be seen as models of cognitive structures and processes in brains, this paper may also be seen to demonstrate the relative strength of the SP system in modelling aspects of human perception and cognition.

\section{Introduction to SP-neural}\label{sp-neural_intro_section}

As we have seen in Section \ref{sp-abstract_outline_section}, SP-abstract is a relatively simple system with descriptive and explanatory power across a wide range of observation and phenomena in artificial intelligence, mainstream computing, mathematics, and human perception and cognition. How can such a system have anything useful to say about the extraordinary complexity of brains and nervous systems, both in their structure and in their workings?

An answer in brief is that SP-neural---a realisation of SP-abstract in terms of neurons, their interconnections, and the transmission of impulses between neurons---may help us to interpret neural structures and processes in terms of the relatively simple concepts in SP-abstract. To the extent that this is successful, it may---like any good theory in any field---help us to understand empirical phenomena in our area of interest, it may help us to make predictions, and it may suggest lines of investigation.

It is anticipated that SP-neural will work in broadly the same way as SP-abstract, but the characteristics of neurons and their interconnections raise some issues that do not arise in SP-abstract and its realisation in the SP computer model. These issues will be discussed at appropriate points in this and subsequent sections.

This section introduces SP-neural in outline, and sections that follow describe aspects of the theory in more detail, drawing where necessary on aspects of SP-abstract that have been omitted from or only sketched in Section \ref{sp-abstract_outline_section}.

\begin{figure}[!htbp]
\centering
\includegraphics[width=0.9\textwidth]{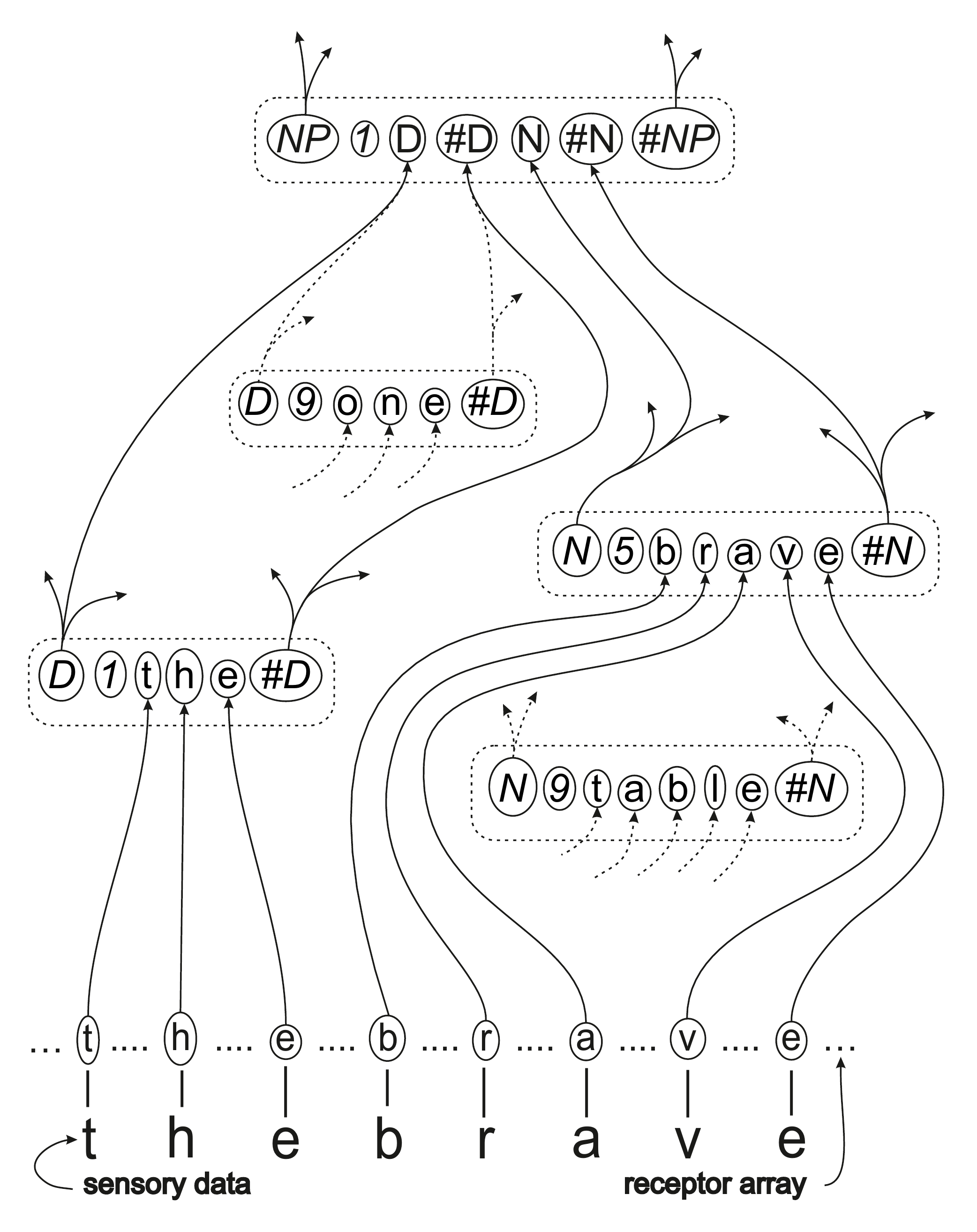}
\caption{A schematic outline of how part of the multiple alignment shown in Figure \ref{fortune_brave_multiple_alignment_figure}, with associated patterns and symbols, may be expressed in SP-neural as neurons and their inter-connections. The meanings of the conventions in the figure, and some complexities that are not shown in the figure, are explained in this main section and ones that follow.}
\label{the_brave_neural_figure}
\end{figure}

Figure \ref{the_brave_neural_figure} shows in outline how a portion of the multiple alignment shown in Figure \ref{fortune_brave_multiple_alignment_figure}, may be realised in SP-neural. with associated patterns and symbols.

\subsection{Sensory data and the receptor array}\label{sp-n_sensory_data_receptor_array_section}

In the figure, `sensory data' at the bottom means the visual, auditory or tactile data entering the system which, in this case, corresponds with the phrase `t h e b r a v e'. In a more realistic illustration, the sensory data would be some kind of analogue signal. Here, the letters are intended to suggest the kinds of low-level perceptual primitives outlined below.

It is envisaged that, with most sensory modalities, the receptor array would be located in the primary sensory cortex. Of course, a lot of processing goes on in the sense organs and elsewhere between the sense organs and the primary sensory cortices. But it seems that most of this early processing is concerned with the identification of the perceptual primitives just mentioned.

As with SP-abstract, it is anticipated that SP-neural will, at some stage, be generalised to accommodate patterns in two dimensions, such as visual images, and then the sensory data may be received in two dimensions, as in the human eye.

Between the sensory data and the {\em receptor array} (above it in the figure), there would be, first, cells that are specialised to receive particular kinds of input (auditory, visual, tactile etc). These send signals to neurons that encode the sensory data as {\em neural symbols}, the neural equivalents of `symbols' in SP-abstract. In the receptor array, each letter enclosed in a solid ellipse represents a neural symbol, expressed as a single neuron or, more likely, a small cluster of neurons. As we shall see (Section \ref{sp-n_encoding_in_receptor_array_section}), the reality is more complex, at least in some cases.

In vision, neural symbols in the receptor array would represent such low-level features as lines, corners, colours, and the like, while in speech perception, they would represent such things as formants, formant ratios and transitions, plosive and fricative sounds, and so on. Whether or how the SP concepts can be applied in the discovery or identification of features like these is an open question \cite[Section 3.3]{sp_extended_overview}. For now, we shall assume that they can be identified and can be used in the creation and use of higher-level structures.

\subsection{Pattern assemblies}\label{sp-n_pattern_assemblies_section}

In the rest of Figure \ref{the_brave_neural_figure}, each broken-line rectangle with rounded corners represents a {\em pattern assembly}---corresponding to a `pattern' in SP-abstract. The word `assembly' has been adopted within the expression `pattern assembly' because the concept is quite similar to Hebb's concept of a `cell assembly'---a cluster of neurons representing a concept or other coherent mental entity. Differences between Hebb's concept of a cell assembly and the SP concept of a pattern assembly are described in Appendix \ref{cell_pattern_assemblies_appendix}.

Within each pattern assembly, as represented in the figure, each character or group of characters enclosed in a solid-line ellipse represents a {\em neural symbol} which, as already mentioned, corresponds to a `symbol' in SP-abstract. As with neural symbols in the receptor array, it is envisaged that each neural symbol would comprise a single neuron or, more likely, a small cluster of neurons.

It is supposed that, within each pattern assembly, there are lateral connections between neural symbols---but these are not shown in the figure.

It is envisaged that most pattern assemblies would represent knowledge that is learned and not inborn, and would be located mainly outside the primary sensory areas of the cortex, in other parts of the sensory cortices. Pattern assemblies that integrate two or more sensory modalities may be located in the `association' areas of the cortex.

Research with fMRI recordings from volunteers \cite{huth_etal_2016} has revealed ``semantic maps'' that ``show that semantic information is represented in rich patterns that are distributed across several broad regions of cortex. Furthermore, each of these regions contains many distinct areas that are selective for particular types of semantic information, such as people, numbers, visual properties, or places. We also found that these cortical maps are quite similar across people, even down to relatively small details.''\footnote{From the website of the Gallant Lab at UC Berkely, retrieved 2016-05-02, \href{http://bit.ly/1WvvLhX}{http://bit.ly/1WvvLhX}. See also ``Brain `atlas' of words revealed'', {\em BBC News}, 2016-04-27, \href{http://bbc.in/1SGESLz}{bbc.in/1SGESLz}.} Of course, this research says nothing about whether or not the knowledge is represented with pattern assemblies and their interconnections. But it does apparently confirm that knowledge is stored in several regions of the cortex and throws light on how it is organised.

Although most parts of the mammalian cerebral cortex has six layers and many convolutions, it may be seen, topologically, as a sheet which is very much broader and wider than it is thick. Correspondingly, it is envisaged that 1D and 2D pattern assemblies will be largely `flat' structures, rather like writing or pictures on a sheet of paper. That said, it is quite possible, indeed likely, that pattern assemblies would take advantage of two or more layers of the cortex, not just one.

Incidentally, since 2D SP patterns may provide a basis for 3D models, as described in \cite[Sections 6.1 and 6.2]{sp_vision}, flat neural structures in the cortex may serve to represent 3D concepts.

\subsection{Connections between pattern assemblies}\label{sp-n_connections_between_pattern_assemblies_section}

In Figure \ref{the_brave_neural_figure}, the solid or broken lines that connect with neural symbols represent axons, with arrows representing the direction of travel of neural impulses. Where two or more connections converge on a neural symbol, we may suppose that, contrary to the simplified way in which the convergence is shown in the figure, there would be a separate dendrite for each connection.

Axons represented with solid lines are ones that would be active when the multiple alignment in Figure \ref{fortune_brave_multiple_alignment_figure} is in the process of being identified. Broken-line connections show a few of the many other possible connections.

As mentioned in Section \ref{sp-n_pattern_assemblies_section}, it is envisaged that there would be one or more neural connections between neighbouring neural symbols within each pattern assembly but these are not marked in the figure.

Compared with what is shown in the figure, it likely that, in reality, there would be more `levels' between basic neural symbols in the receptor array and ID-neural-symbols representing pattern assemblies for relatively complex entities like the words `one', `brave', `the', and `table', as shown in the figure. In this connection, it is perhaps worth emphasising that, as with the modelling of hierarchical structures in multiple alignments (Section \ref{sp-a_multiple_alignment_section}), while pattern assemblies may form `strict' hierarchies, this is not an essential feature of the concept, and it is likely that many neural structures formed from pattern assemblies may be only loosely hierarchical or not hierarchical at all.

\subsection{SP-neural, quantities of knowledge, and the size of the brain}

Given the foregoing account of how knowledge may be represented in the brain, a question that arises is ``Are there enough neurons in the brain to store what a typical person knows?'' This is a difficult question to answer with any precision but an attempt at an answer, described in \cite[Section 11.4.9]{wolff_2006}, reaches the tentative conclusion that there are. In brief:

\begin{itemize}

  \item Given that estimates of the size of the human brain range from $10^{10}$ up to $10^{11}$ neurons,\footnote{This is consistent with another estimate, not quoted in \cite[Section 11.4.9]{wolff_2006}, that there may be as many as 86 billion neurons in the human brain \cite{herculano-houzel_2012}.} we may estimate, via calculations given in \cite[Section 11.4.9]{wolff_2006}, that the `raw' storage capacity of the brain is between approximately 1000 MB and 10,000 MB.

  \item Given a conservative estimate that, using SP compression mechanisms, compression by a factor of 3 may be achieved across all kinds of knowledge, our estimates of the storage capacity of the brain will range from about 3000 MB up to about 30,000 MB.

  \item Assuming: 1) That the average person knows only a relatively small proportion of what is contained in the {\em Encyclopaedia Britannica} (EB); 2) That the average person knows lots of `everyday' things that are {\em not} in the EB; 3) That the `everyday' things that we {\em do} know are roughly equal to the things in the EB that we {\em do not} know; Then (4), we may conclude that the size of the EB provides a rough estimate of the volume of information that the average person knows.

  \item The EB can be stored on two CDs in compressed form. Assuming that most of the space is filled, this equates to 1300 MB of compressed information or approximately 4000 MB of information in uncompressed form.

  \item This 4000 MB estimate of what the average person knows is the same order of magnitude as our range of estimates (3000 MB to 30,000 MB) of what the human brain can store.

  \item Even if the brain stores two or three copies of its compressed knowledge---to guard against the risk of losing it, or to speed up processing, or both---our estimate of what needs to be stored (lets say $4000 \times 3 = 12,000$ MB) is still within the 3000 MB to 30,000 MB range of estimates of what the brain can store.

\end{itemize}

\subsection{Neural processing}\label{sp-n_neural_processing_section}

In broad terms, it is envisaged that, for a task like the parsing of natural language or pattern recognition:

\begin{enumerate}

    \item SP-neural will work firstly by receiving sensory data and interpreting it as neural symbols in the receptor array---with excitation of the neural symbols that have been identified.

        Excitatory signals would be sent from those excited neural symbols to pattern assemblies that can receive signals from them directly. In Figure \ref{the_brave_neural_figure}, these would be all the pattern assemblies except the topmost pattern assembly.

        Within each pattern assembly, excitatory signals will spread laterally via the connections between neighbouring neural symbols.

        Pattern assemblies would become excited, roughly in proportion to the number of excitatory signals they receive.

    \item At this stage, there would be a process of selecting amongst pattern assemblies to identify one or two that are most excited.

    \item From those pattern assemblies---more specifically, the neural ID-symbols at the beginnings and ends of those pattern assemblies---excitatory signals would be sent onwards to other pattern assemblies that may receive them. In Figure \ref{the_brave_neural_figure}, this would be the topmost pattern assembly (that would be reached immediately after the first pass through stages 2 and 3).

        As in stage 1, the level of excitation of any pattern assembly would depend on the number of excitatory signals it receives, but building up from stage to stage so that the highest-level pattern assemblies are likely to be most excited.

    \item Repeat stages 2 and 3 until there are no more pattern assemblies that can be sent excitatory signals.

\end{enumerate}

The `winning' pattern assembly or pattern assemblies, together with the structures below them that have, directly or indirectly, sent excitatory signals to them, may be seen as neural analogues of multiple alignments (NAMAs), and we may guess that they provide the best interpretations of a given portion of the sensory data.

If the whole sentence, `\texttt{f o r t u n e f a v o u r s t h e b r a v e}', is processed by SP-neural with pattern assemblies that are analogues of the SP patterns provided for the example shown in Figure \ref{fortune_brave_multiple_alignment_figure}, we may anticipate that the overall result would be a pattern of neural excitation that is an analogue of the multiple alignment shown in that figure.

When a neural symbol or pattern assembly has been `recognised' by participating in a winning (neural) multiple alignment, we may suppose that some biochemical or physiological aspect of that structure is increased as an at least approximate measure of the frequency of occurrence of the structure, in accordance with the way in which SP-abstract keeps track of the frequency of occurrence of symbols and patterns (Section \ref{sp-a_early_learning_section}).

Some further possibilities are discussed in Sections \ref{sp-n_more_detail_section} and \ref{sp-n_inhibition_section}.

\section{Some more detail}\label{sp-n_more_detail_section}

The bare-bones description of SP-neural in Section \ref{sp-neural_intro_section} is probably inaccurate in some respects and is certainly too simple to work effectively. This section and the ones that follow describe some other features which are likely to figure in a mature version of SP-neural, drawing on relevant empirical evidence where it is available.

\subsection{Encoding of information in the receptor array}\label{sp-n_encoding_in_receptor_array_section}

With regard to the encoding of information in the receptor array, it seems that the main possibilities are these:

\begin{enumerate}

  \item {\em Explicit alternatives}. For the receptor array to work as described in Section \ref{sp-neural_intro_section}, it should be possible to encode sensory inputs with an `alphabet' of alternative values at each location in the array, in much the same way that each binary digit (bit) in a conventional computer may be set to have the value 0 or 1, or how a typist may enter any one of an alphabet of characters at any one location on the page. At each location in the receptor array, each option may be provided in the form of a neuron or small cluster of neurons. Here, there seem to be two main options:

      \begin{enumerate}

        \item {\em Horizontal distribution of alternatives}. The several alternatives may be distributed `horizontally', in a plane that is parallel to the surface of the cortex.

        \item {\em Vertical distribution of alternatives}. The several alternatives may be distributed `vertically' between the outer and inner surfaces of the cortex, and perpendicular to those surfaces.

      \end{enumerate}

  \item {\em Implicit alternatives}. At each location there may be a neuron or small cluster of neurons that, via some kind of biochemical or neurophysiological process, may be `set' to any one of the alphabet of alternative values.

  \item {\em Rate codes}. Something like the intensity of a stimulus may be encoded via ``an interaction between [the] firing rates and the number of neurons [that are] activated by [the] stimulus.'' \cite[p.~503]{squire_etal_2013}.

  \item {\em Temporal codes}. A stimulus that varies with time may be encoded via ``the time-varying pattern of activity in small groups of receptors and central neurons.'' ({\em ibid.}).

\end{enumerate}

In support of option 1.a, there is evidence that neurons in the visual cortex (of cats) are arranged in columns perpendicular to the surface of the cortex, where, for example, all the neurons in a given column respond most strongly to a line at one particular angle in the field of view, that---within a `hypercolumn' containing several columns---the preferred angle increases progressively from column to column, and that there are many hypercolumns across the primary visual cortex \cite{barlow_1982}. ``Hubel and Wiesel point out that the organization their results reveal means that each small region, about $1 mm^2$ at the surface, contains a complete sequence of ocular dominance and a complete sequence of orientation preference.'' \cite[pp.~148--149]{barlow_1982}.

\begin{figure}[!htbp]
\centering
\includegraphics[width=0.8\textwidth]{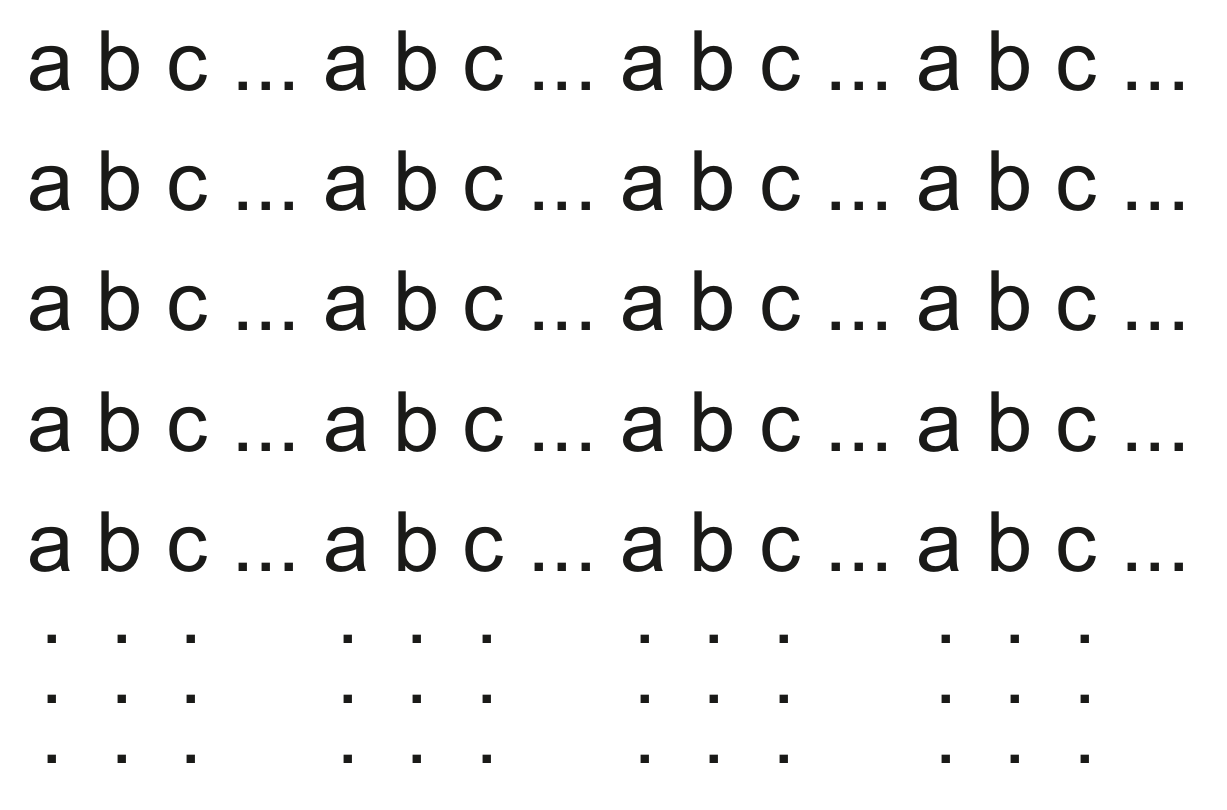}
\caption{Schematic representation of one hypercolumn in the receptor array in the cortex. Each letter represents a neural symbol that responds to a particular small pattern in the sensory data. The ellipsis, `...', in each row and each column represents other neural symbols that would be shown in a more comprehensive representation of the given hypercolumn. Each vertical sequence of letters, all of one kind such as `\texttt{a}' or `\texttt{b}' represents a simple column in the cortex.}
\label{receptor_array_detail_figure}
\end{figure}

Leaving out the results for ocular dominance, these observations are summarised schematically in Figure \ref{receptor_array_detail_figure}. In terms of this scheme, the way in which the receptor array is shown in Figure \ref{the_brave_neural_figure}, is a considerable simplification---each neural symbol in the receptor array in that figure should really be replaced by a hypercolumn.

With something like the intensity of a stimulus, it seems that, at least in some cases: ``...~activity in one particular population of somatosensory neurons ...~leads the CNS to interpret it as painful stimulus ....'' \cite[p.~503]{squire_etal_2013}, while ``An entirely separate population of neurons ...~would signal light pressure.'' ({\em ibid.}). Since it is likely that relevant receptors appear repeatedly across one's skin, this appears to be another example of option 1.a.

There seems to be little evidence of encoding via option 1.b. Indeed, since the concept of a cortical column is, in effect, defined by the fact that all the neurons in any one column have the same kind of receptive field, this seems to rule out the 1.b option (see also Section \ref{why_multiple_neurons_section}).

But, with respect to option 2, it appears that in some cases, as noted above, the intensity of a stimulus may be encoded via the rates of firing of neurons, together with the numbers of neurons that are activated (option 3). And, since we can perceive and remember time-varying stimuli such as the stroking of a finger across one's skin, or the rising or falling pitch of a note, some kind of temporal encoding must be available (option 4).

Here, it must be acknowledged that options 3 and 4 appear superficially to be outside the scope of the SP theory, in view of the emphasis in many examples on discrete atomic symbols. But, as we know from the success of digital recording, or indeed digital computing, any continuum may be encoded digitally, in keeping with the digital nature of the SP theory. How the SP theory may be applied to the digital encoding and processing of continua has been discussed elsewhere in relation to vision \cite{sp_vision} and the development of autonomous robots \cite{sp_autonomous_robots}.

\subsection{Why are there multiple neurons with the same receptive fields in columns in the cortex?}\label{why_multiple_neurons_section}

As we have seen (Section \ref{sp-n_encoding_in_receptor_array_section}), some aspects of vision are mediated via columns of neurons in the primary visual cortex in which each column contains many neurons with receptive fields that are all the same, all of them responding, for example, to a line in the visual field with a particular orientation.

Why, at each of several locations across the visual cortex, should there be many neurons with the same receptive field, not just one? There seem to be two possible answers to this question (and they are not necessarily mutually exclusive):

\begin{itemize}

    \item {\em Encoding of sensory patterns}. If, in the receptor array, we wish to encode two or more patterns such as `\texttt{m e t}' and `\texttt{h e m}', they need to be independent of each other, with repetition of the `\texttt{e}' neural symbol, otherwise there will be the possibly unwanted implication that such things as `\texttt{m e m}' or `\texttt{h e t}' are valid patterns.

    \item {\em Error-reducing redundancy}. At any given location in the receptor array, multiple instances of a given neural symbol may help to guard against the problems that may arise if there is only neural symbol at that location and if, for any reason, it becomes partially or fully disabled.

\end{itemize}

With regard to the first point, the receptor array may have a useful role to play, {\em inter alia}, as a short-term memory for many sensory patterns pending their longer-term storage (Section \ref{sp-n_speed_expressiveness_section}). In vision, for example, the receptor array may store many short glimpses of a scene, as outlined in Section \ref{sp-n_we_see_less_than_we_think_section}, until such time as further processing may be applied to weld the many glimpses into a coherent structure ({\em ibid.}) and to transfer that structure to longer-term memory.

\subsection{The labelled line principle}\label{sp-n_labelled_line_principle_section}

Section \ref{sp-n_neural_processing_section} suggests that normally, at some early stage in sensory processing, raw sensory data is encoded in terms of the excitation of neuronal symbols in a receptor array, then excited neural symbols send excitatory signals to appropriate neural symbols within pattern assemblies, and pattern assemblies that are sufficiently excited send excitatory signals on to other pattern assemblies, and so on. As we shall see (Section \ref{sp-n_inhibition_section}), it is likely that, in this processing, there will also be a role for inhibitory processes.

At first sight, it may be thought that, in the same way that each location in the receptor array should provide an alphabet of alternative encodings (Section \ref{sp-n_encoding_in_receptor_array_section}), the same should be true for the location of each neural symbol within each pattern assembly. But if a neural symbol in a pattern assembly (let's call it `NS1') receives signals only from neural symbols in the receptor array that represent a given feature, let us say, `a', then, in accordance with the `labelled line' principle \cite[p.~503]{squire_etal_2013}, NS1 also represents `a'.

For most sensory modalities, this principle applies all the way from each sense organ, through the thalamus, to the corresponding part of the primary sensory cortex.\footnote{Thus, for example, ``Even within one function, mappings of neurons [within the thalamus] are preserved so that there is separation of neurons providing touch information from the arm versus from the leg and of neurons responding to low versus high sound frequencies ....'' \cite[p.~507]{squire_etal_2013}. Also, ``Nuclei in the central pathways often contain multiple maps.'' but ``The functional significance of multiple maps in general, however, remains to be clarified.'' ({\em ibid.}).} It seems reasonable to suppose that the same principle will apply onwards from each primary sensory cortex into non-primary sensory cortices and non-sensory association areas.

\subsection{How the ordering or 2D arrangement of neural symbols may be respected}

In SP-neural, as in SP-abstract and the SP computer model, the process of matching one pattern with another should respect the orderings of symbols. For example, `\texttt{A B C D}' matched with `\texttt{A B C D}' should be rated more highly in terms of information compression than, for example, `\texttt{A B C D}' matched with `\texttt{C A D B}'.\footnote{A possible exception is when one pattern is a mirror image or inversion of another, since Leonardo da Vinci, by repute, could read mirror writing as easily as ordinary writing, and it is now well established that people wearing inverting spectacles can learn quite quickly to see the world as if it was the right way up \cite{stratton_1897}.}

It appears that this problem may be solved by the adoption, within SP-neural, of the following feature of natural sensory systems:

\begin{quote}

    ``Receptors within [the retina and body surface] communicate with ganglion cells and those ganglion cells with central neurons in a strictly ordered fashion, such that relationships with neighbours are maintained throughout. This type of pattern, in which neurons positioned side by side in one region communicate with neurons positioned side-by-side in the next region, is called a {\em typographic pattern}.'' \cite[p.~504]{squire_etal_2013} (emphasis in the original).

\end{quote}

\subsection{How to accommodate the variable sizes of sensory patterns}

A prominent feature of human visual perception is that we can recognise any given entity over a wide range of viewing distances, with correspondingly wide variations in the size, on the retina, of the image of that entity.

For any model of human visual perception that is based on a simplistic or naive process for the matching of patterns, this aspect of visual perception would be hard to reproduce or to explain. But the SP system is different: 1) Knowledge of entities that we may recognise are always stored in a compressed form; 2) The process of recognition is a process of compressing the incoming data; 3) The overall effect is that an image of a thing to be recognised can be matched with stored knowledge of that entity, regardless of the original size of the image.

As an example, consider how the concept of an equilateral triangle (as white space bounded by three black lines all of the same length) may be stored and how an image of such a triangle may be recognised. Regarding storage, there are three main redundancies in any image of that kind of triangle: 1) The white space in the middle may be seen as repeated instances of the symbol `white'; 2) Each of the three sides of the triangle may be seen as repeated instances of the symbol `black' or `point'; and 3) There is redundancy in that the three sides of the triangle are the same.

\begin{figure}[!htbp]
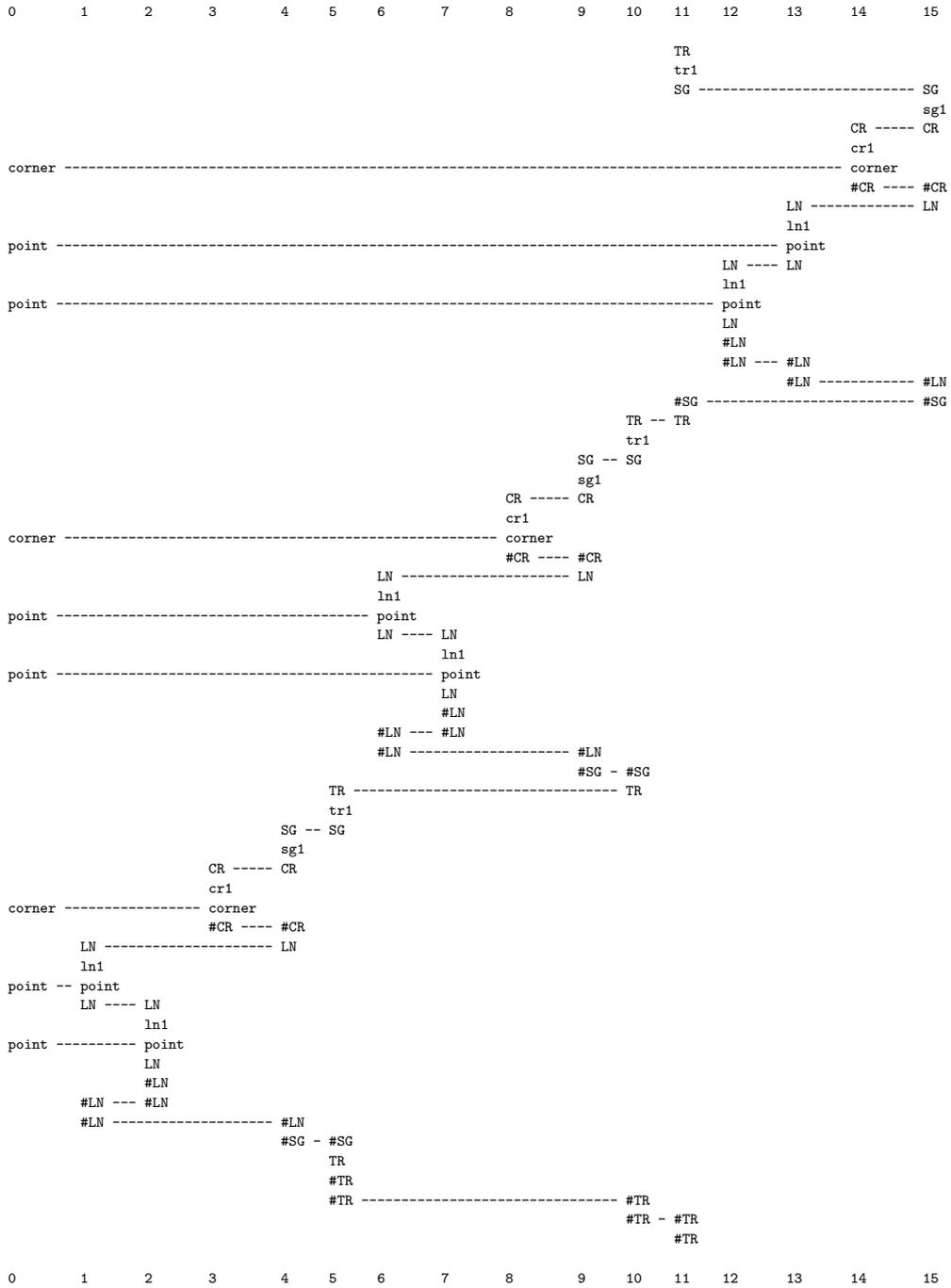

\fontsize{06.50pt}{07.80pt}
\centering
{\bf
\begin{BVerbatim}
0        1       2       3        4     5     6       7       8        9     10    11    12      13      14       15

                                                                                   TR
                                                                                   tr1
                                                                                   SG --------------------------- SG
                                                                                                                  sg1
                                                                                                         CR ----- CR
                                                                                                         cr1
corner ------------------------------------------------------------------------------------------------- corner
                                                                                                         #CR ---- #CR
                                                                                                 LN ------------- LN
                                                                                                 ln1
point ------------------------------------------------------------------------------------------ point
                                                                                         LN ---- LN
                                                                                         ln1
point ---------------------------------------------------------------------------------- point
                                                                                         LN
                                                                                         #LN
                                                                                         #LN --- #LN
                                                                                                 #LN ------------ #LN
                                                                                   #SG -------------------------- #SG
                                                                             TR -- TR
                                                                             tr1
                                                                       SG -- SG
                                                                       sg1
                                                              CR ----- CR
                                                              cr1
corner ------------------------------------------------------ corner
                                                              #CR ---- #CR
                                              LN --------------------- LN
                                              ln1
point --------------------------------------- point
                                              LN ---- LN
                                                      ln1
point ----------------------------------------------- point
                                                      LN
                                                      #LN
                                              #LN --- #LN
                                              #LN -------------------- #LN
                                                                       #SG - #SG
                                        TR --------------------------------- TR
                                        tr1
                                  SG -- SG
                                  sg1
                         CR ----- CR
                         cr1
corner ----------------- corner
                         #CR ---- #CR
         LN --------------------- LN
         ln1
point -- point
         LN ---- LN
                 ln1
point ---------- point
                 LN
                 #LN
         #LN --- #LN
         #LN -------------------- #LN
                                  #SG - #SG
                                        TR
                                        #TR
                                        #TR -------------------------------- #TR
                                                                             #TR - #TR
                                                                                   #TR

0        1       2       3        4     5     6       7       8        9     10    11    12      13      14       15
\end{BVerbatim}
}
\caption{A multiple alignment produced by the SP computer model showing how a one-dimensional analogue of how an equilateral triangle may be perceived, as described in the text. Adapted from \cite[Figure 8]{sp_alternatives}, with permission.}
\label{triangle_multiple_alignment_figure}
\end{figure}

All three sources of redundancy may be encoded recursively as suggested in Figure \ref{triangle_multiple_alignment_figure}, which shows a multiple alignment modelling the recognition of a one-dimensional analogue of a triangle.

Column 0 shows information about the triangle to be recognised, comprising three `\texttt{corners}' and three sides of the triangle, each one represented by just two `\texttt{points}'.

The pattern `\texttt{LN ln1 point LN \#LN \#LN}' in columns 1 and 2 is a self-referential and thus recursive definition of a line as a sequence of `\texttt{points}'. It is self-referential because, within the body of the pattern, it contains a reference to itself via the symbols at the beginning and end of the pattern: `\texttt{LN \#LN}'. Because there is no limit to this recursion, it may represent a line containing any number of points. In a similar way, a second side is encoded via the same pattern in columns 6 and 7, and, again with the same pattern, the third line is encoded in columns 12 and 12.

In columns 4, 9 and 15 in the figure, the pattern `\texttt{SG sg1 CR \#CR LN \#LN \#SG}' shows one of the three elements of a triangle as a corner (`\texttt{CR \#CR}') followed by a line (`\texttt{LN \#LN}'). And the recursion to encode multiple instances of that structure is in self-referential occurrences of the pattern `\texttt{TR tr1 SG \#SG TR \#TR \#TR}' in columns 5, 10 and 22. Strictly speaking, the encoding is for a polygon, not a triangle, because there is nothing to stop the recursive repetition of `\texttt{SG sg1 CR \#CR LN \#LN \#SG}'. And, in terms of the problem, as described above, the representation is incomplete because there is nothing to show that the three sides of the triangle are the same.

These encodings account for the redundancy in the repetition of points along a line and also the redundancy in the repetition of three sides of a triangle. In a 2D version, they would also account for the redundancy in the white space within the body of the triangle, because they would allow most of the white space to be eliminated via shrinkage of the representation to the minimum needed to express the concept of a triangle.

\subsection{We see much less than we think we see}\label{sp-n_we_see_less_than_we_think_section}

Most people with normal vision have a powerful sense that their eyes are a window on to a kind of cinema screen that shows what we are looking at with great detail from left to right and from top to bottom. But research shows otherwise:

\begin{itemize}

    \item In the phenomenon of {\em inattentional blindness}, people may fail to notice salient things in their visual fields when they are looking for something else, even if they are trained observers. In a recent demonstration \cite{drew_atal_2013}, radiologists were asked to search for lung-nodules in chest x-rays but many of them (83\%) failed to notice the image of a gorilla, 48 times the size of the average nodule, that was inserted into one of the radiographs.

    \item In the phenomenon of {\em change blindness}, people often fail to notice large changes to visual scenes. For example, if a conversation between two people---the investigator and the experimental subject---is interrupted by a door being carried between them, the experimental subject may fail to notice, when the door has gone by, that the person they are speaking to is different from the person they were speaking to before \cite{simons_ambinder_2005}.

    \item Although each of our eyes has a blind spot\footnote{See ``Blind spot (vision)'', {\em Wikipedia}, \href{http://bit.ly/1oI0vyI}{bit.ly/1oI0vyI}, retrieved 2016-04-08.}, we don't notice it, even when we are viewing things with one eye. Apparently, our brains interpolate what is likely to be in the blind part of our visual field.

\end{itemize}

It seems that part of the reason for this failure to see things is that photoreceptors are concentrated at the fovea \cite[p.~502]{squire_etal_2013}, and cones are only found in that region ({\em ibid.}), so that, with two eyes, we are, to a large extent, looking at the world through a keyhole composed of two circumscribed and largely overlapping views, one from each eye.

It seems that our sense that the world is displayed to us on a wide and deep cinema screen is partly because our perception of any given scene draws heavily on our memories of similar scenes and partly because we can piece together what will normally be a partial view of what we are looking at from many short glimpses through the `keyhole' as we move our gaze around the scene.

The SP theory provides an interpretation for these things as follows:

\begin{itemize}

    \item The theory provides an account in some detail of how New (sensory) information may be related to Old (stored) information and how an interpretation of the New information may be built up via the creation of multiple alignments.

    \item The theory provides an account of how we can piece together a picture of something, or indeed a 3D model of something, from many small but partially-overlapping views, in much the same way that: 1) With digital photography, it is possible to create a panoramic picture from several partially-overlapping images; 2) The views in Google's Streetview are built up from many partially-overlapping pictures; 3) A 3D digital image of an object may be created from partially-overlapping images of the object, taken from viewpoints around it. These things are discussed in \cite[Sections 5.4 and 6.1]{sp_vision}.

\end{itemize}

With regard to the second point, it should perhaps be said that partial overlap between `keyhole' views is not an essential part of building up a big picture from smaller views. But if two or more views do overlap, it is useful if they can be stitched together, thus removing the overlap. And partial overlap may be helpful in establishing the relative positions of two or more views.

\subsection{A resolution problem and its possible resolution}\label{sp-n_resolution_problem_section}

As we have seen (Section \ref{sp-n_encoding_in_receptor_array_section}), each hypercolumn in the primary visual cortex of cats occupies about $1 mm^2$ at the surface of the cortex, and it seems likely that each such hypercolumn provides a means of encoding one out of an alphabet of perceptual primitives, such as a line at a particular angle.

Assuming that this interpretation is correct, and if we view the primary visual cortex as if it was film in an old-style camera or the image sensor in a digital camera, it may seem that the encoding of perceptual primitives, with $1 mm^2$ for each one, is remarkably crude. How could such a system---with the area of the primary visual cortex corresponding to the area of our field of view---create that powerful sense that, through our eyes, we see a detailed `cinema screen' view of the world (Section \ref{sp-n_we_see_less_than_we_think_section}).

Part of the answer is probably that we see much less than we think we see (Section \ref{sp-n_we_see_less_than_we_think_section}). But it seems likely that another part of the answer is to reject the assumption that the area of the primary visual cortex corresponding to the area of our field of view. In the light of the remarks in Section \ref{sp-n_we_see_less_than_we_think_section}, it seems likely that, normally, in each of the previously-mentioned glimpses of a scene, most of the primary visual cortex is applied in the assimilation and processing of information capture by the fovea and, perhaps, parts of the retina that are very close to the fovea.\footnote{In support of this idea: ``{\bf Cortical magnification} describes how many neurons in an area of the visual cortex are `responsible' for processing a stimulus of a given size, as a function of visual field location. In the center of the visual field, corresponding to the fovea of the retina, a very large number of neurons process information from a small region of the visual field. If the same stimulus is seen in the periphery of the visual field (i.e. away from the center), it would be processed by a much smaller number of neurons. The reduction of the number of neurons per visual field area from foveal to peripheral representations is achieved in several steps along the visual pathway, starting already in the retina \cite{barghout-stein_1999}.'', {\em Wikipedia}, \href{http://bit.ly/1qJsQX1}{bit.ly/1qJsQX1}, emphasis in the original, retrieved 2016-04-14.} In that case, what appears superficially to be a rather course-grained recording and analysis of visual data, may actually be very much more detailed. As described in Section \ref{sp-n_we_see_less_than_we_think_section}, it seems likely that our view of any scene is built up partly from memories and partly from many small snapshots or glimpses of the scene.

\subsection{Grandmother cells, localist and distributed representations}\label{sp-n_grandmother_cells_section}

In terms of concepts that have been debated about how knowledge may be represented in the brain, the ID-neural-symbols for any pattern assembly are very much like the concept of a {\em grandmother cell}---a cell or small cluster of cells in one's brain that represents one's grandmother so that, if the cell or cells were to be lost, one would lose the ability to recognise one's grandmother.

It seems that the weight of observational and experimental evidence favours the belief that such cells do exist \cite{gross_2002}. This is consistent with the observation that people who have suffered a stroke or are suffering from dementia may lose the ability to recognise members of their close family.

Since SP-neural, like Hebb's \cite{hebb_1949} theory of cell assemblies, proposes that concepts are represented by coherent groups of neurons in the brain, it is very much a `localist' type of theory. As such, it is quite distinct from `distributed' types of theory that propose that concepts are encoded in widely-distributed configurations of neurons, without any identifiable location or centre.

However, just to confuse matters, SP-neural does {\em not} propose that all one's knowledge about one's grandmother would reside in a pattern assembly for that lady. Probably, any such pattern assembly would, in the manner of object-oriented design as discussed in Section \ref{sp-n_non-syntactic_knowledge_section} and illustrated in Figure \ref{class_hierarchy_figure}, be connected to and inherit features from a pattern assembly representing grandmothers in general, and from more general pattern assemblies such as pattern assemblies for such concepts as `person' and `woman'. And again, a pattern assembly for `person' would not be the sole repository of all one's knowledge about people. That pattern assembly would, in effect, contain `references' to pattern assemblies describing the parts of a person, their physiology, their social and political life, and so on.

Thus, while SP-neural is unambiguously localist, it proposes that knowledge of any entity or concept is likely to be encoded not merely in one pattern assembly for that entity or concept but also in many other pattern assemblies in many parts of the cortex, and perhaps elsewhere.

\subsection{Positional invariance}\label{sp-n_positional_invariance_section}

With something simple like a touch on the skin, or a pin prick, it is not too difficult to see how the sensation may be transmitted to the brain via any one of many relevant receptors located in many different areas of the skin. But with something more complex, like an image on the retina of a table, a house, or a tree, and so on, it is less straightforward to understand how we might recognise such a thing in any part of our visual field.

For each entity to be recognised, it seems necessary at first sight to provide connections, directly or indirectly, from every part of the receptor array to the relevant pattern assembly. In terms of the schematic representation shown in Figure \ref{the_brave_neural_figure}, it would mean repeating the connections for `\texttt{t h e}' and `\texttt{b r a v e}' in each of many parts of the receptor array. Bearing in mind the very large number of different things we may recognise, the number of necessary connections would become very large, perhaps prohibitively so.

However, things may be considerably simplified via either or both of two provisions:

\begin{enumerate}

    \item For reasons outlined in Section \ref{sp-n_we_see_less_than_we_think_section}, it seems likely that, with vision, we build up our perception of a scene, partly from memories of similar scenes and partly via many relatively narrow `keyhole' views of what is in front of us. If that is correct, and if, as suggested in Section \ref{sp-n_resolution_problem_section}, most of the primary visual cortex is devoted to analysing information received via the fovea and, perhaps, via parts of the retina that are very close to the fovea, then the need to provide for any given pattern in many parts of the receptor array may be greatly reduced. Since, by moving our eyes, we may view any part of a scene, it is possible that any given entity would need only one or two sets of connections between the receptor array and the pattern assembly for that entity.

    \item As noted in Section \ref{sp-n_connections_between_pattern_assemblies_section}, it seems likely that, with regard to Figure \ref{the_brave_neural_figure}, there would, in a more realistic example, be several levels of structure between neural symbols in the receptor array and relatively complex structures like words. At the first level above the receptor array there would be pattern assemblies for relatively small recurrent structures, and the variety of such structures would be relatively small. This should ease any possible problems in connecting the receptor array to pattern assemblies.

\end{enumerate}

If it turns out that the number of necessary connections is indeed too large to be practical, or if there is empirical evidence against such numbers, then a possible alternative to what has been sketched in this paper is some kind of dynamic system for the making and breaking of connections between the receptor array and pattern assemblies. It seems likely that permanent or semi-permanent connections would be very much more efficient and the balance of probabilities seems to favour such a scheme.

In connection with positional invariance, it is relevant to note that ``...~lack of localization is quite common in higher-level neurons: receptive fields become larger as the features they represent become increasingly complex. Thus, for instance, neurons that respond to faces typically have receptive fields that cover most of the visual space. For these cells, large receptive fields have a distinct advantage: the preferred stimulus can be identified no matter where it is located on the retina.'' \cite[p.~579]{squire_etal_2013}. A tentative and partial explanation of this observation is that repetition of neurons that are sensitive to each of several categories of low-level feature---in the receptor array and as ID-neural-symbols for `low-level' pattern assemblies---is what allows positional invariance to develop at higher levels.

\section{Non-syntactic knowledge in SP-neural}\label{sp-n_non-syntactic_knowledge_section}

As was emphasised in Section \ref{sp-abstract_outline_section}, the SP system (SP-abstract) has strengths and potential in the representation and processing of several different kinds of knowledge, not just the syntax of natural language. That versatility has been achieved using the mechanisms in SP-abstract that were outlined in that section. If those mechanisms can be modelled in SP-neural, it seems likely that the several kinds of knowledge that may be represented and processed in SP-abstract may also be represented and processed in SP-neural.

\begin{figure}[!htbp]
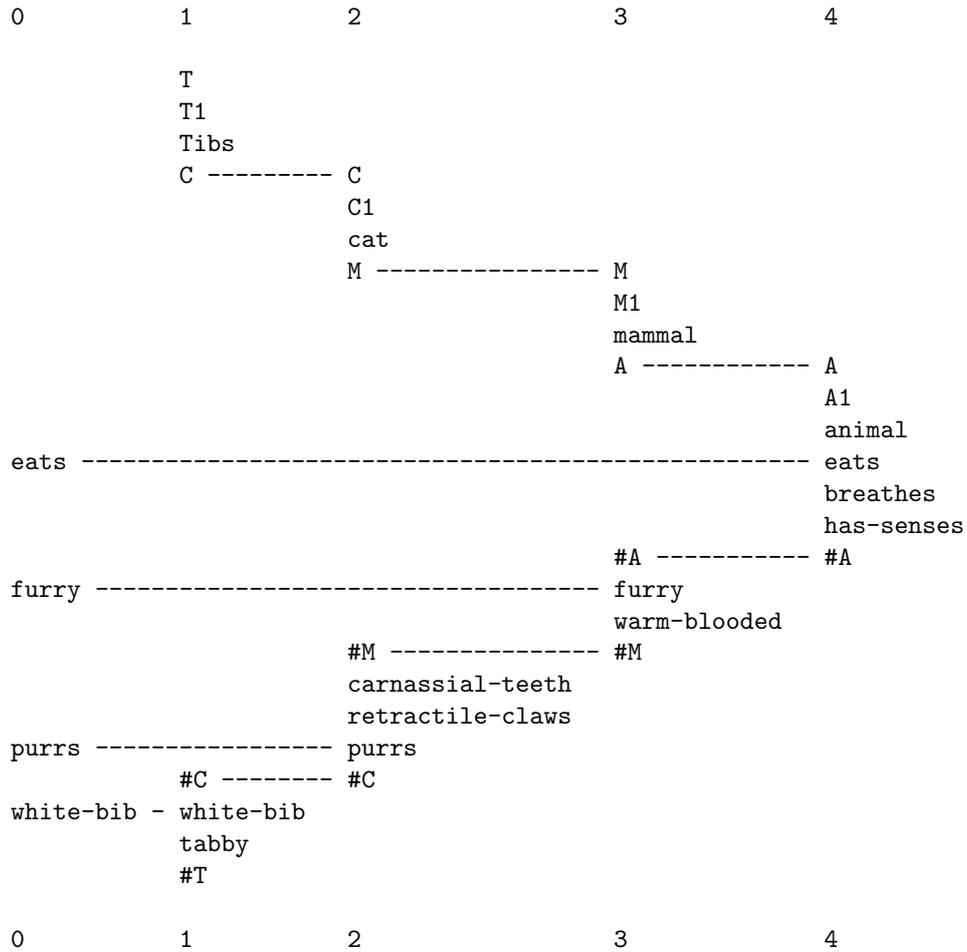

\fontsize{10.00pt}{12.00pt}
\centering
{\bf
\begin{BVerbatim}
0           1           2                  3              4

            T
            T1
            Tibs
            C --------- C
                        C1
                        cat
                        M ---------------- M
                                           M1
                                           mammal
                                           A ------------ A
                                                          A1
                                                          animal
eats ---------------------------------------------------- eats
                                                          breathes
                                                          has-senses
                                           #A ----------- #A
furry ------------------------------------ furry
                                           warm-blooded
                        #M --------------- #M
                        carnassial-teeth
                        retractile-claws
purrs ----------------- purrs
            #C -------- #C
white-bib - white-bib
            tabby
            #T

0           1           2                  3              4
\end{BVerbatim}
}
\caption{The best multiple alignment found by the SP computer model with four one-symbol New patterns representing attributes of an unknown creature and a collection of Old patterns representing different creatures and classes of creature.}
\label{class_hierarchy_figure}
\end{figure}

\begin{figure}[!htbp]
\centering
\includegraphics[width=0.9\textwidth]{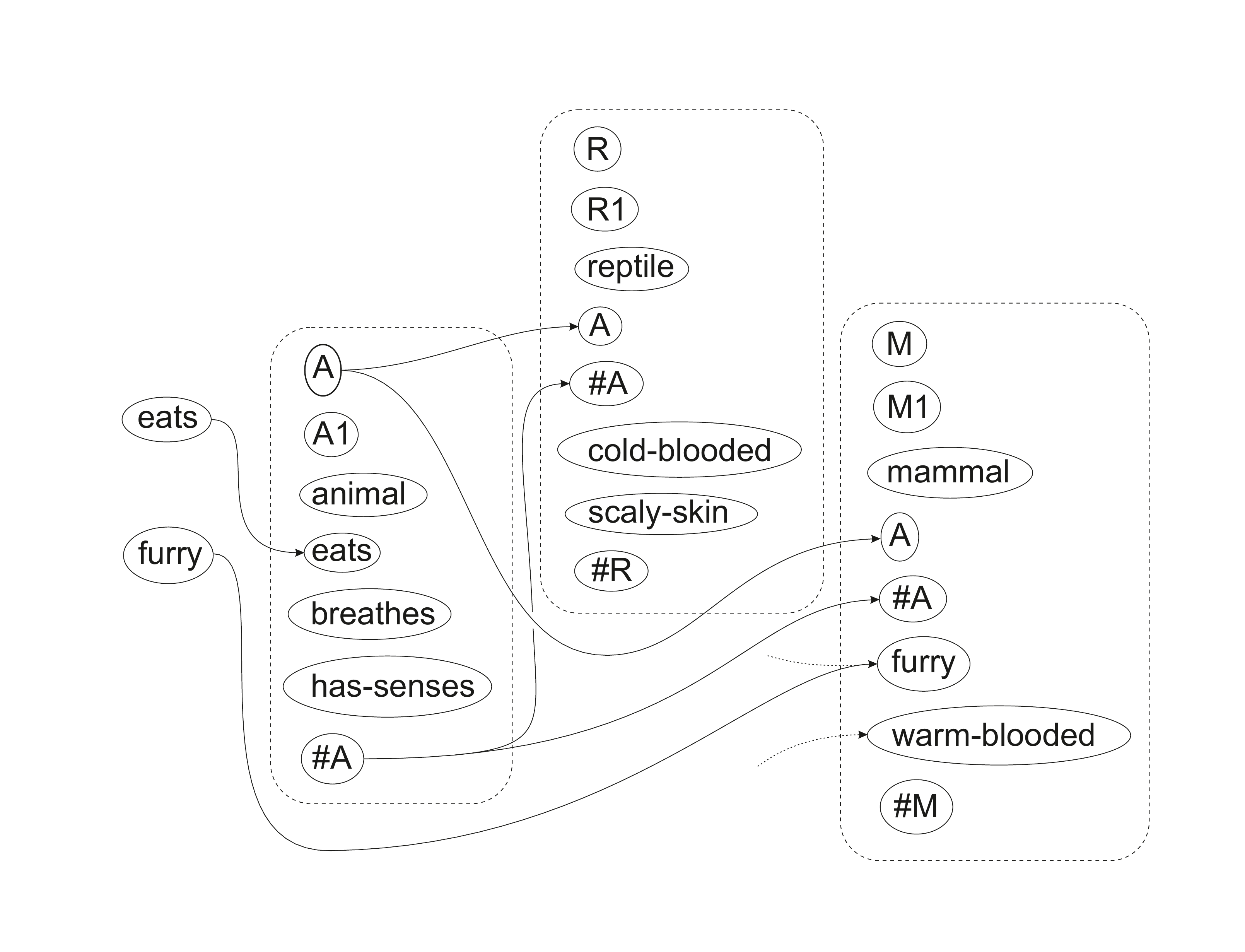}
\caption{How part of the multiple alignment shown in Figure \ref{class_hierarchy_figure} may be realised in SP-neural---showing two of the attributes from column 0 in the multiple alignment and with `animal' and `mammal' pattern assemblies corresponding to patterns from columns 4 and 3---with an associated pattern assembly for `reptile'. The conventions are the same as in Figure \ref{class_hierarchy_figure}.}
\label{class_hierarchy_neural_figure}
\end{figure}

As an illustration, Figure \ref{class_hierarchy_figure}\footnote{Compared with the multiple alignments shown in Figures \ref{partial_match_figure} and \ref{fortune_brave_multiple_alignment_figure}, this multiple alignment has been rotated by $90\degree$. The choice between these alternative presentations of multiple alignments depends entirely on what fits best on the page.} shows a simple example of how, via multiple alignment, the SP computer model may recognise an unknown creature at several different levels of abstraction, and Figure \ref{class_hierarchy_neural_figure} suggests how part of the multiple alignment, with associated patterns, may be realised in terms of pattern assemblies and their inter-connections.

Figure \ref{class_hierarchy_figure} shows the best multiple alignment found by the SP computer model with four symbols representing attributes of an unknown creature (shown in column 0) and a collection of Old patterns representing different creatures and classes of creature, some of which are shown in columns 1 to 4, one pattern per column. In a more detailed and realistic example, symbols like `\texttt{eats}', `\texttt{retractile-claws}', and `\texttt{breathes}', would be represented as patterns, each with its own structure.

From this multiple alignment, we can see that the unknown creature has been identified as an animal (column 4), as a mammal (column 3), as a cat (column 2) and as a specific cat, `Tibs' (column 1). It is just an accident of how the SP computer model has worked in this case that the order of the patterns across columns 1 to 4 of the multiple alignment corresponds with the level of abstraction of the classifications. In general, the order of patterns in columns above 0 is entirely arbitrary, with no significance.

Figure \ref{class_hierarchy_neural_figure} shows how part of the multiple alignment from Figure \ref{class_hierarchy_figure} may be realised in SP-neural. The figure contains pattern assemblies for `animal' and `mammal', corresponding to patterns from columns 4 and 3 of the multiple alignment. Notice that the left-right order of the pattern assemblies is different from the order of the patterns in the multiple alignment, in accordance with the remarks, above, about the workings of the SP computer model, and also because there is no reason to believe that pattern assemblies are represented in any particular order.

Neural connections amongst the things that have been mentioned so far are very much the same as alignments between neural symbols in Figure \ref{class_hierarchy_figure}: `eats' on the left connects with `eats' in the `animal' pattern assembly; `furry' connects with `furry' in the `mammal' pattern assembly, and the `A' and `\#A' connections for those two pattern assemblies correspond with the alignments of symbols in the multiple alignment. As in Figure \ref{the_brave_neural_figure}, some neural connections are shown with broken lines to suggest that they would be relatively inactive during the neural processing which identifies one or more `good' NAMAs. And as before, it is envisaged that there would be one or more neural connections between each neural symbol and its immediate neighbours within each pattern assembly, but these are not marked in the figure.

The inclusion of a pattern assembly for `reptile' in Figure \ref{class_hierarchy_neural_figure}, with some of its neural connections, is intended to suggest some of the processing involved in identifying one or more winning NAMAs. In the same way that the pattern for `mammal' is receiving excitatory signals from the pattern for `animal', one would expect excitatory signals to flow to pattern assemblies for the other main groups of animals, including reptiles. Ultimately, `reptile' would fail to feature in any winning NAMA because of evidence from the neural symbols `furry', `purrs' and `white-bib'.

\section{Repetition and recursion}\label{repetition_recursion_section}

Like any good database or dictionary, the repository of Old patterns in SP-abstract should only contain one copy of any given SP pattern. But in something like {\em Jack Sprat could eat no fat, His wife could eat no lean}, the words {\em could}, {\em eat}, and {\em no} each occur twice. With an example like this, it seems reasonable to suppose that there is only one stored pattern for each of the repeated words, and likewise for the many other examples of entities that are repeated within something larger, witness the many legs of a centipede.

In SP-abstract, this apparent difficulty has been overcome by saying that each SP pattern in a multiple alignment is an {\em appearance} of the pattern, not the pattern itself---which allows us to have multiple instances of a pattern in a multiple alignment without breaking the rule that the repository of Old patterns should contain only one copy of each pattern. But in SP-neural, it is not obvious how to create an `appearance' of a pattern assembly that is not also a physical structure of neurons and their interconnections---but the speed with which we can understand natural language seems to rule out what appears to be the relatively slow growth of new neurons and their interconnections.

How we can create new mental structures quickly arises again in other connections, as discussed in Section \ref{sp-n_speed_expressiveness_section}. If we duck these questions for the time being and return to parsing, it may be argued that with something like {\em Jack Sprat could eat no fat, His wife could eat no lean}, the first instance of {\em could} is represented only for the duration of the word by the stored pattern for {\em could}, so that the same pattern can be used again to represent the second instance of {\em could}---and likewise for {\em eat} and {\em no}. But it appears that this line of reasoning does not work with a recursive structure like {\em the very very very fast car}.

\begin{figure}[!htbp]
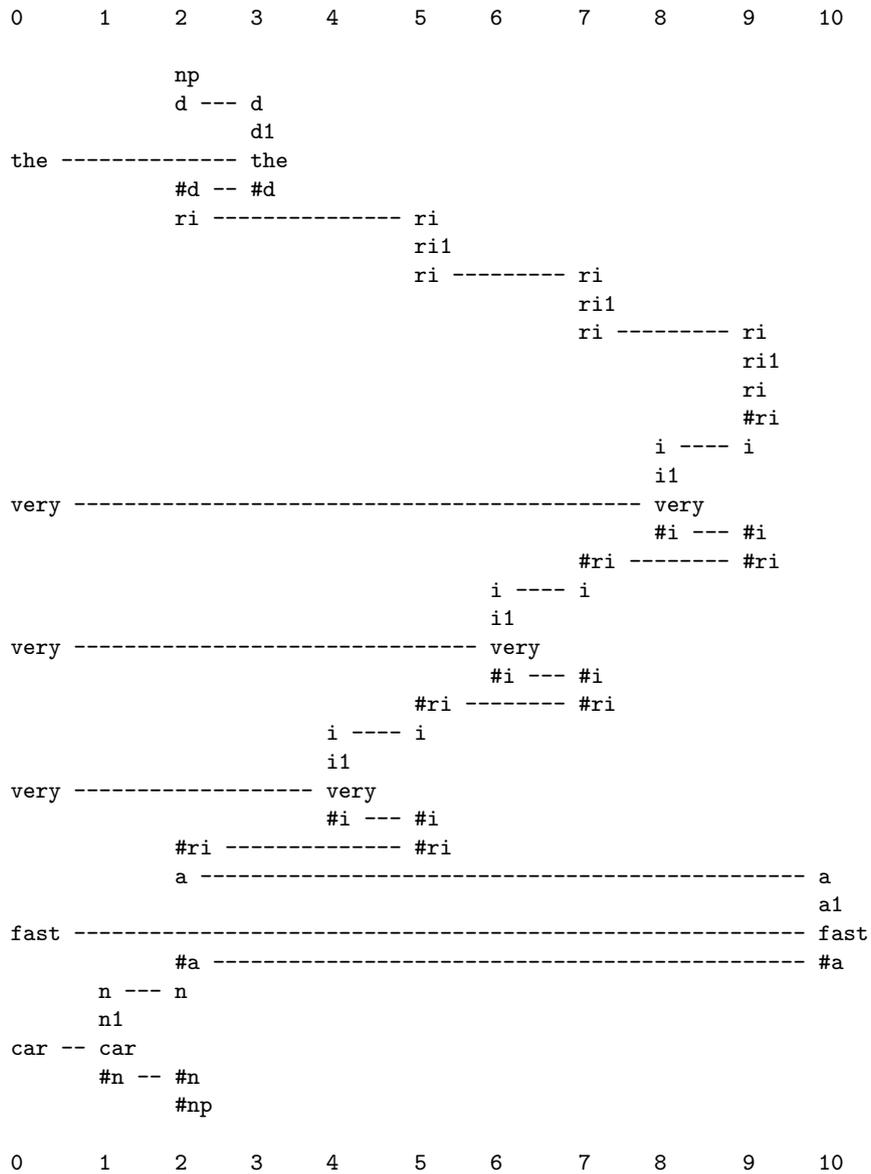

\fontsize{09.00pt}{10.80pt}
\centering
{\bf
\begin{BVerbatim}
0      1     2     3     4      5     6      7     8      9     10

             np
             d --- d
                   d1
the -------------- the
             #d -- #d
             ri --------------- ri
                                ri1
                                ri --------- ri
                                             ri1
                                             ri --------- ri
                                                          ri1
                                                          ri
                                                          #ri
                                                   i ---- i
                                                   i1
very --------------------------------------------- very
                                                   #i --- #i
                                             #ri -------- #ri
                                      i ---- i
                                      i1
very -------------------------------- very
                                      #i --- #i
                                #ri -------- #ri
                         i ---- i
                         i1
very ------------------- very
                         #i --- #i
             #ri -------------- #ri
             a ------------------------------------------------ a
                                                                a1
fast ---------------------------------------------------------- fast
             #a ----------------------------------------------- #a
       n --- n
       n1
car -- car
       #n -- #n
             #np

0      1     2     3     4      5     6      7     8      9     10
\end{BVerbatim}
}
\caption{A multiple alignment produced by the SP computer model showing how recursion may be modelled in the SP system.}
\label{ma_recursion_figure} 
\end{figure}

Native speakers of English know that with a phrase like {\em the very very very fast car}, the word {\em very} may in principle be repeated any number of times. This observation, coupled with the observation that recursive structures are widespread in English and other natural languages, suggests strongly that the most appropriate parsing of the phrase is something like the multiple alignment shown in Figure \ref{ma_recursion_figure}. Here, the repetition of {\em very} is represented via three appearances of the pattern `\texttt{ri ri1 ri \#ri i \#i \#ri}', a pattern which is self-referential because the inner pair of symbols `\texttt{ri \#ri}' can be matched with the same two symbols, one at the beginning of the pattern and one at the end. Because the recursion depends on at least two instances of `\texttt{ri ri1 ri \#ri i \#i \#ri}' being `live' at the same time, it seems necessary for SP-neural to be able to model multiple appearances of any pattern.

That conclusion, coupled with the above-mentioned arguments from the speed at which we can speak, and the speed with which we can imagine new things, argues strongly that SP-neural---and any other neural theory of cognition---must have a means of creating new mental structures quickly. It seems unlikely that these things could be done via the growth of new neurons and their interconnections.

The tentative answer suggested here is that, in processes like parsing or pattern recognition, including examples with recursion like that shown in Figure \ref{ma_recursion_figure}, virtual copies of pattern assemblies may be created and destroyed very quickly via the switching on and switching off of synapses (see Section \ref{sp-n_speed_expressiveness_section}). Clearly, more detail is needed for a fully satisfactory answer.

\begin{figure}[!htbp]
\centering
\includegraphics[width=0.9\textwidth]{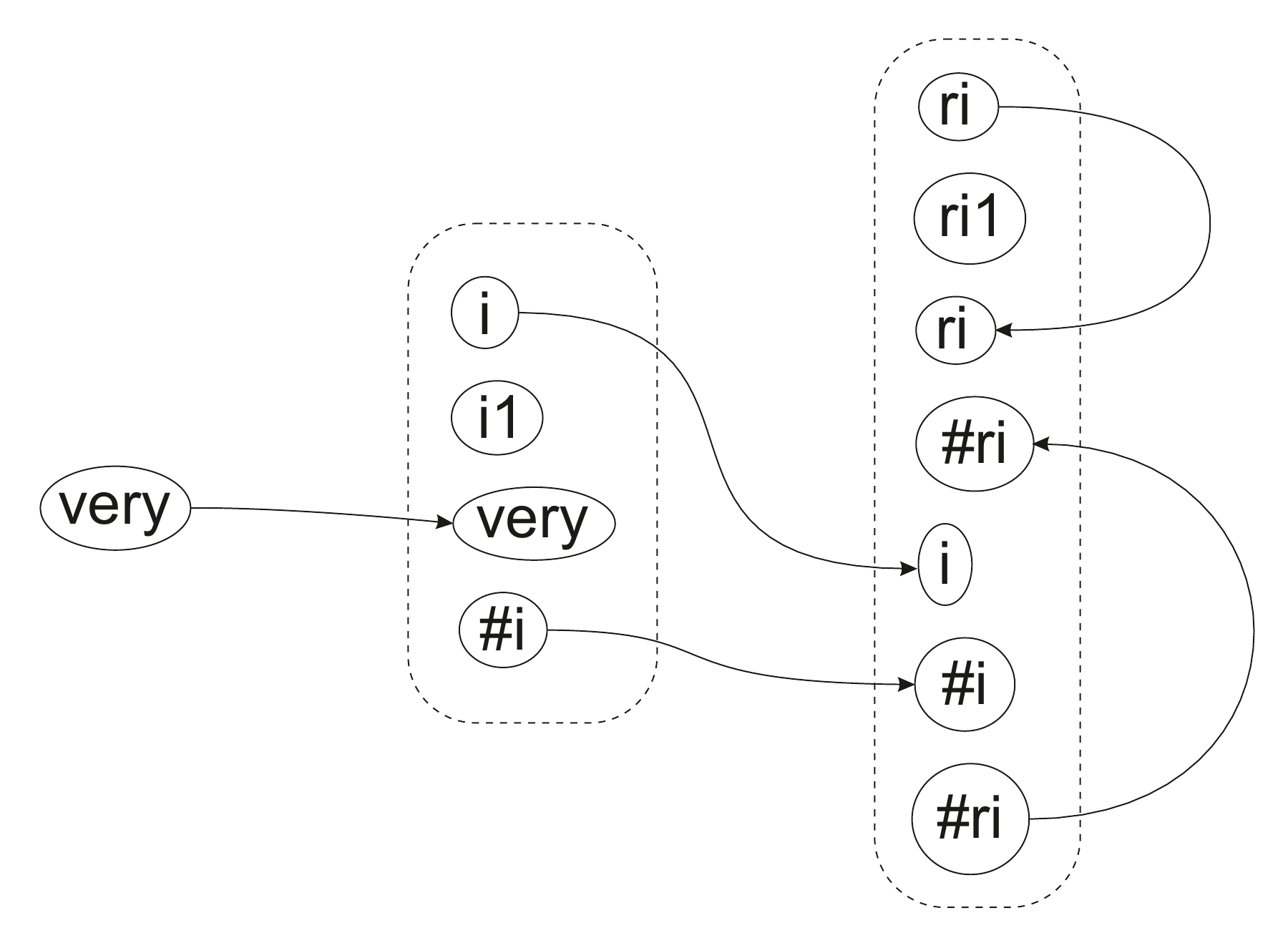}
\caption{A schematic example showing how recursive structures may be modelled in SP-neural.}
\label{neural_recursion_figure}
\end{figure}

Pending that better answer, Figure \ref{neural_recursion_figure} shows tentatively how recursion may be modelled in SP-neural, with neural symbols and pattern assemblies corresponding to selected symbols and patterns in Figure \ref{ma_recursion_figure}. On the left of that figure, we can see how the neural symbol `\texttt{very}' connects with a matching neural symbol in the pattern assembly `\texttt{i i1 very \#i}'. Further right, we can see how the first and last neural symbols in `\texttt{i i1 very \#i}' connect with matching neural symbols in the pattern assembly `\texttt{ri ri1 ri \#ri i \#i \#ri}'.

In the figure, the self-referential nature of the pattern assembly `\texttt{ri ri1 ri \#ri i \#i \#ri}' can be seen in the neural connection between `\texttt{ri}' at the beginning of that pattern assembly and the matching neural symbol in the body of the same pattern assembly, and likewise for `\texttt{\#ri}' at the end of the pattern assembly. Although it is unclear how this recursion may achieve the effect of repeated appearances of the pattern assembly at the speed with which we understand or produce speech, the analysis appears to be sounder than what is described in \cite[Section 11.4.2]{wolff_2006}, especially Figure 11.10 in that section.

\section{SP-neural: an output perspective}\label{sp-n_output_section}

\sloppy An inspection of Figure \ref{the_brave_neural_figure}---showing how, in SP-neural, a small portion of natural language may be analysed by pattern assemblies and their interconnections---may suggest that if we wish to reverse the process---to create language instead of analysing it---then the innervation would need to be reversed: we may guess that two-way neural connections would be needed to support the production of speech or writing as well as their interpretation.

But a neat feature of SP-abstract is that one set of Old patterns, together with the processes for building multiple alignments, will support both the analysis and the production of language. So it is reasonable to suppose that if SP-neural works at all, similar duality will apply to pattern assemblies and their interconnections, without the need for two-way connections amongst pattern assemblies and neural symbols (but see Section \ref{sp-n_efferent_projections_section}).

Of course, speaking or writing would need peripheral motor processes that are different from the peripheral sensory processes required for listening or reading, but, more centrally, the processes for analysing language or producing it may use the same mechanisms.\footnote{Of course, things are a little more complicated with output processes because sensory feedback is normally an important part of speaking or writing. But the central point remains that, peripherally, speaking is not the same as listening to speech, and likewise for writing and reading.}

\begin{figure}[!htbp]
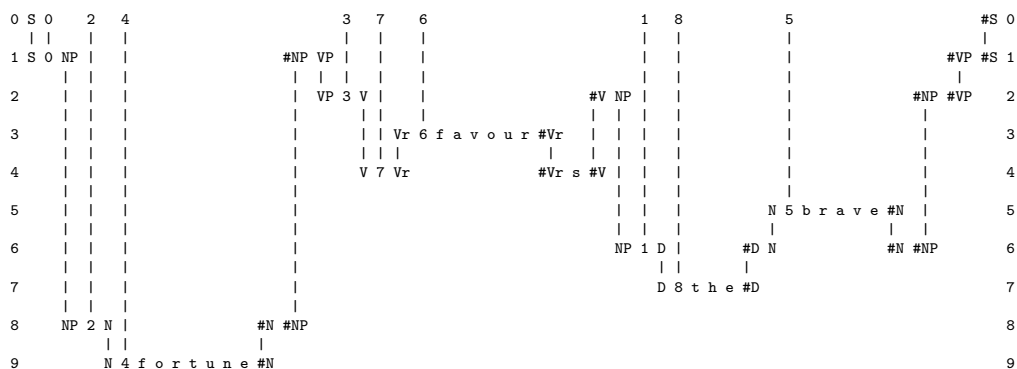

\fontsize{06.00pt}{07.20pt}
\centering
{\bf
\begin{BVerbatim}
0 S 0    2   4                         3   7    6                         1   8            5                      #S 0
  | |    |   |                         |   |    |                         |   |            |                      |
1 S 0 NP |   |                  #NP VP |   |    |                         |   |            |                  #VP #S 1
      |  |   |                   |  |  |   |    |                         |   |            |                   |
2     |  |   |                   |  VP 3 V |    |                   #V NP |   |            |              #NP #VP    2
      |  |   |                   |       | |    |                   |  |  |   |            |               |
3     |  |   |                   |       | | Vr 6 f a v o u r #Vr   |  |  |   |            |               |         3
      |  |   |                   |       | | |                 |    |  |  |   |            |               |
4     |  |   |                   |       V 7 Vr               #Vr s #V |  |   |            |               |         4
      |  |   |                   |                                     |  |   |            |               |
5     |  |   |                   |                                     |  |   |          N 5 b r a v e #N  |         5
      |  |   |                   |                                     |  |   |          |             |   |
6     |  |   |                   |                                     NP 1 D |       #D N             #N #NP        6
      |  |   |                   |                                          | |       |
7     |  |   |                   |                                          D 8 t h e #D                             7
      |  |   |                   |
8     NP 2 N |               #N #NP                                                                                  8
           | |               |
9          N 4 f o r t u n e #N                                                                                      9
\end{BVerbatim}
}
\caption{The best multiple alignment produced by the SP computer model with the same Old patterns as for the multiple alignment shown in Figure \ref{fortune_brave_multiple_alignment_figure} but with the New pattern comprising an appropriate sequence of ID-symbols, `\texttt{S 0 2 4 3 7 6 1 8 5 \#S}', as described in the text.}
\label{output_figure}
\end{figure}

The reason that SP-abstract, as expressed in the SP computer model, can work in `reverse' so to speak, is that, from a multiple alignment like the one shown in Figure \ref{fortune_brave_multiple_alignment_figure}, a code pattern like `\texttt{S 0 2 4 3 7 6 1 8 5 \#S}' may be derived, as outlined in Section \ref{sp-a_deriving_code_pattern_section}. Then, if that code pattern is presented to the SP system as a New pattern, the system can recreate the original sentence, `\texttt{f o r t u n e f a v o u r s t h e b r a v e}', as shown in Figure \ref{output_figure}.

It is likely that, in a more fully-developed account, code patterns would represent connections between syntax and semantics so that they may provide the means of generating sentences from meanings. As noted in Section \ref{sp-a_evaluation_systems_section}, preliminary trials show how, with the SP computer model, sentences may be generated from meanings and {\em vice versa}.

\subsection{An answer to the apparent paradox of ``decompression by compression''}

That the SP system should be able to reconstruct a sentence that was originally compressed by means of the same system (Section \ref{sp-n_output_section}) may seem paradoxical. How is it that a system that is dedicated to information compression should be able, so to speak, to drive compression in reverse?

A resolution of this apparent paradox is described in \cite[Section 3.8]{wolff_2006}. In brief, the key to the conjuring trick is to ensure that, after the sentence has been compressed, there is enough residual redundancy in the code pattern to allow further compression, and to ensure that this further compression will achieve the effect of reconstructing the sentence.

\subsection{Meanings in the analysis and production of language}

Of course, parsing a sentence (as shown in Section \ref{sp-a_multiple_alignment_section}) or constructing a sentence from a code pattern (as shown in Section \ref{sp-n_output_section}) are very artificial applications with natural language. Normally, when we read some text or listen to someone speaking, we aim to derive meaning from the writing or the speech. And when we write or speak, it seems, intuitively, that the patterns of words that we are creating are derived from some kind of underlying meaning that we are trying to express.

It is envisaged that, in future development of SP-abstract and the SP computer model, the ID-symbols in code patterns will provide some kind of bridge between syntactic forms and representations of meanings, thus facilitating the processes of understanding the meanings of written or spoken sentences and of creating sentences to express particular meanings.

As noted at the end of Section \ref{sp-a_evaluation_systems_section}, there are preliminary examples of how, with the SP computer model, a sentence may be analysed for its meaning \cite[Section 5.7, Figure 5.18]{wolff_2006}, and how the same sentence may be derived from a representation of its meaning {\em ibid.}, Figure 5.19).

\subsection{But there are projections from the sensory cortex to subcortical nuclei}\label{sp-n_efferent_projections_section}

Although as we have seen earlier in Section \ref{sp-n_output_section}, SP-neural, via principles established in SP-abstract, provides for the creation of language, and other kinds of knowledge, without the need for efferent connections from the cortex back along the path of afferent nerves, there is evidence that such connections do exist:

\begin{quote}

    ``Neurons of the cerebral cortex send axons to subcortical regions .... Subcortical projections are to those nuclei in the thalamus and brainstem that provide ascending sensory information. By far the most prominent of these is to the thalamus: the neurons of a primary sensory cortex project back to the same thalamic nucleus that provides input to the cortex. This system of descending connections is truly impressive because the number of descending corticothalamic axons greatly exceeds the number of ascending thalamocortical axons. These connections permit a particular sensory cortex to control the activity of the very neurons that relay information to it.'' \cite[p.~509]{squire_etal_2013}.

\end{quote}

But the descending nerves described in this quotation may have a function that is quite different from the creation of sentences or other patterns of activity. One possible role for such nerves may be ``the focussing of activity so that relay neurons most activated by a sensory stimulus are more strongly driven and those in surrounding less well activated regions are further suppressed.'' \cite[p.~509]{squire_etal_2013}.

\section{Possible roles for inhibition in SP-neural}\label{sp-n_inhibition_section}

A familiar observation is that, if something like a fan is switched on near us, we notice the noise for a while and then come to ignore it. And if, later, the fan is switched off, we notice the relative quiet for a while and then cease to be aware of it. In general, it seems that we are relatively sensitive to changes in our environment and relatively insensitive to things that remain constant.

It has been accepted for some time that the way we adapt to constant stimuli is due to inhibitory neural structures and processes in our brains and nervous systems, that inhibitory structures and processes are widespread in the animal kingdom, and that they have a role in reducing the amount of information that we need to process \cite{von_bekesy_1967}.

Regarding the last point, it is clearly inefficient for anyone to be constantly registering, second-by-second, the noise of a nearby fan: `\texttt{noise, noise, noise, noise, noise, ...}' and likewise for the state of relative quietness when the fan is switched off. In terms of information theory, there is {\em redundancy} in the second-by-second recurrence of the noise (or quietness), and we can eliminate most or all of the redundancy---and thus compress the information---by simply recording that the noise is `on' and that it is continuing (and likewise, {\em mutatis mutandis}, for quiet). This is the `run-length encoding' technique for compression of information,\footnote{See ``Run-length encoding'', {\em Wikipedia}, \href{http://bit.ly/21JlB1T}{bit.ly/21JlB1T}, retrieved 2016-03-04.} it is essentially what adaptation does, and, in neural tissue, it appears to be mediated largely by `lateral' inhibition.

With lateral inhibition in sensory neurons, there are inhibitory connections between neighbouring neurons so that, when they are both stimulated, they tend to inhibit each other, and thus reduce their rates of firing where there is strong uniform stimulation. But inhibition is reduced where strong stimulation gives way to weaker stimulation, leading to a local swing in the rate of firing (\cite{ratliff_etal_1963}; see also \cite[Section 2.3.1]{wolff_2006}; there is more about lateral inhibition in \cite[p.~505]{squire_etal_2013}). There are similar effects in the time dimension. Again, Barlow \cite{barlow_1982} says, in connection with neurons in the mammalian cortex that receive inputs from both eyes, ``...~it is now clear that input from one eye can, and frequently does, inhibit the effects of input from the other eye,~...'' (p.~147).

Taking these observations together, we may abstract a general rule: {\em When, in neural processing, two or more signals are the same, they tend to inhibit each other, and when they are different, they don't.} The overall effect should be to detect redundancy in information and to reduce it, whilst retaining non-redundant information, in accordance with the central principle in the SP theory---that much of computing and cognition may, to a large extent, be understood as information compression.

In a similar vein: ``Lateral inhibition represents the classic example of a general principle: most neurons in sensory systems are best adapted for detecting changes in the external environment.~...~As a rule, it is change which has the most significance for an animal ...~This principle can also be explained in terms of information processing. Given a world that is filled with constants---with uniform objects, with objects that move only rarely---it is most efficient to respond only to changes.'' \cite[p.~578]{squire_etal_2013}.

In view of the widespread occurrence of inhibitory mechanisms in the brain,\footnote{``These [aspiny or sparsely spiny nonpyramida] interneurons constitute approximately 15 to 30\% of the total population of cortical neurons, and they appear to be mostly GABAergic, representing the main components of inhibitory cortical circuits ....'' \cite[p.~45]{squire_etal_2013}; ``Synaptic inhibition in the mammalian brain is mediated principally by GABA receptors.'' ({\em ibid.}, p.~169); ``One of the great mysteries of synaptic integration is why there are so many different types of inhibitory interneurons.~...~more than 20 different types of inhibitory interneuron have been described in the CA1 region of the hippocampus alone.'' ({\em ibid.}, p.~249).} and in view of their apparent importance for the compression of information, and thus for selective advantage \cite[Section 4]{sp_foundations}, it is pertinent to ask what role or roles they may play in SP-neural. Here are some possibilities:

\begin{itemize}

  \item {\em Low-level sensory features}. At relatively `low' levels in sensory processing, it appears that, as noted above, lateral inhibition has a role in identifying such things as boundaries between uniform areas, meaning lines. It may also have a role in identifying other kinds of low-level sensory features mentioned in Section \ref{sp-n_sensory_data_receptor_array_section}.

  \item {\em Information compression via the matching and unification of patterns (ICMUP)}. As noted in Section \ref{elements_of_sp-a_section}, SP-abstract, and the SP computer model, is founded on the principle that information compression may be achieved by the matching and unification of patterns (ICMUP). Here, there appear to be two possible roles for inhibition:

      \begin{itemize}

        \item As we have seen, lateral inhibition can have the effect of inhibiting signals from neighbouring sensory neurons when they are receiving stimulation which is the same of nearly so. This may be seen as an example of ICMUP.

        \item In accordance with the rule suggested above, inhibitory processes may serve as a means of detecting redundancy between a New pattern assembly like `\texttt{t a b l e}' and an Old pattern assembly like `\texttt{N 9 t a b l e \#N}':

            \begin{itemize}

              \item We may suppose that there are inhibitory links between neighbouring neural symbols in the Old pattern assembly so that, if all of the neural symbols in the body of the pattern assembly (ie, `\texttt{t a b l e}') are stimulated, or most of them, then mutual inhibition amongst those neural symbols will suppress their response. And, as with lateral inhibition in sensory neural tissue, inhibition in one area can mean enhanced responses at the boundaries with neighbouring areas, which, in this case, would be the ID-symbols `\texttt{N}' and `\texttt{9}' on the left, and `\texttt{\#N}' on the right. Then, excitatory signals from `\texttt{N}' and `\texttt{\#N}' may go on to higher-level patterns that contain nouns, as suggested by the broken-line links from those two neural symbols in Figure \ref{the_brave_neural_figure}. Since there is no link to export excitatory signals from `\texttt{9}', no such signals would be sent.

              \item Alternatively, we may suppose that a stored pattern assembly like `\texttt{N 9 t a b l e \#N}' has a background rate of firing and that, when matching stimulation is received for the neural symbols `\texttt{t a b l e}', the background rate of firing in the corresponding neural symbols in `\texttt{N 9 t a b l e \#N}' is reduced, with an associated upswing in the rates of firing of the neural symbols `\texttt{N}' and `\texttt{9}' and `\texttt{\#N}', as before.

            \end{itemize}

      \end{itemize}

  \item {\em Sharpening choices amongst alternatives}. As mentioned in Section \ref{sp-n_neural_processing_section}, the process of forming neural analogues of multiple alignments (NAMAs) means identifying one or two of the most excited pattern assemblies, with structures below them that feed excitation to them. Here, inhibition may play a part by enhancing the status of the most excited pattern assemblies and suppressing the rest. How inhibition may achieve that kind of effect is discussed quite fully by von B{\'e}k{\'e}sy \cite[Chapters II and V]{von_bekesy_1967}, and also in \cite{shamma_1985}.

\end{itemize}

More information and discussion about the possible roles of inhibition in the cerebral cortex may be found in \cite{isaacson_scanziani_2011}.

\section{Unsupervised learning in SP-neural}\label{sp-n_unsupervised_learning_section}


This section considers how the learning processes in SP-abstract, which are outlined in Sections \ref{sp-a_early_learning_section} and \ref{sp-a_later_learning_section}, may be realised in SP-neural.

It seems likely that neural structures for the detection of `low level' features like lines and corners in vision, or formant ratios and transitions in hearing, are largely inborn,\footnote{``For all systems except the olfactory, the receptor neurons you were born with are the ones you will live with.'' \cite[p.~503]{squire_etal_2013}} although ``It is a curious paradox that, while [Hubel and Wiesel] have consistently argued for a high degree of ontogenetic determination of structure and function in the visual system, they are also the authors of the best example of plasticity in response to changed visual experience.'' \cite[p.~150]{barlow_1982}, and ``It has ...~been shown convincingly that the orientation preference of cells can be modified, ...'' ({\em ibid.}). Also, ``In the somatosensory system, if input from a restricted area of the body surface is removed by severing a nerve or by amputation of a digit, the portion of the cortex that was previously responsive to that region of the body surface becomes responsive to neighbouring regions ....'' \cite[p.~508]{squire_etal_2013}.

But it is clear that most of what we learn in life is at a `higher' level which, in SP-neural, will be acquired via the the creation and destruction of pattern assemblies, as discussed in the following subsections.

\subsection{Creating Old pattern assemblies}\label{sp-n_creating_old_pattern_assemblies_section}

Let us suppose that a young child hears the speech equivalent of `\texttt{t h e b i g h o u s e}' in accordance with the example in Section \ref{sp-a_early_learning_section}. As we have seen, when the repository of Old patterns is empty or nearly so, New patterns are stored directly as Old patterns, somewhat like a recording machine, but with the addition of ID-symbols at their beginnings and ends.

It seems unlikely that a young child would grow new neurons to store a newly-created Old pattern assembly like `\texttt{A 1 t h e b i g h o u s e \#A}', as discussed in Section \ref{sp-a_early_learning_section}. It seems much more likely that a pattern assembly like that would be created by some kind of modification of pre-existing neural tissue comprising sequences or areas of unassigned neural symbols with lateral connections between them as suggested in Section \ref{sp-n_pattern_assemblies_section}. Pattern assemblies would be created by the switching on and off of synapses at appropriate points, in a manner that is more like a tailor cutting up pre-woven cloth than someone knitting or crocheting each item from scratch.

In accordance with the labelled line principle (Section \ref{sp-n_labelled_line_principle_section}), the meaning of each symbol in a newly-created pattern assembly would be determined by what it is connected to, as described in Section \ref{sp-n_creating_neural_connections_section}.

Similar principles would apply when Old patterns are created from partial matches between patterns, as described in Section \ref{sp-a_early_learning_section}.

\subsection{Creating connections between pattern assemblies}\label{sp-n_creating_neural_connections_section}

As with the laying down of newly-created Old patterns (Section \ref{sp-n_creating_old_pattern_assemblies_section}), it seems unlikely that connections between pattern assemblies, like those shown in Figure \ref{the_brave_neural_figure}, would be created by growing new axons or dendrites. It seems much more likely that such connections would be established by switching on synapses between each of the two neurons to be connected and pre-existing axons or dendrites, somewhat like the making of connections in a telephone exchange (see Section \ref{sp-n_speed_expressiveness_section}).

This idea, together with the suggestions in Section \ref{sp-n_creating_old_pattern_assemblies_section} about how Old pattern assemblies may be created, is somewhat like the way in which an `uncommitted logic array' (ULA)\footnote{See `Gate array', {\em Wikipedia}, \href{http://bit.ly/1UdB46j}{bit.ly/1UdB46j}, retrieved 2016-03-20.} may, via small modifications, be made to function like any one of a wide variety of `application-specific integrated circuits' (ASICs),\footnote{See `Application-specific integrated circuit', {\em Wikipedia}, \href{http://bit.ly/1pUs2y8}{bit.ly/1pUs2y8}, retrieved 2016-03-20.} or how a `field-programmable gate array' (FPGA)\footnote{See `Field-programmable gate array', {\em Wikipedia}, \href{http://bit.ly/1Hgi9iH}{bit.ly/1Hgi9iH}, retrieved 2016-03-20.} may be programmed to function like any one of a wide variety of integrated circuits.

\subsection{Destruction of pattern assemblies and their interconnections}\label{sp-n_destroying_neural_connections_section}

In the SP theory, patterns and pattern assemblies are never modified---they are either created or destroyed. The latter process occurs mainly in the process of searching for `good' grammars to describe a given set of New patterns, as outlined in Section \ref{sp-a_later_learning_section}. At each stage, when a few `good' grammars are retained in the system, the rest are discarded. This means that any pattern assembly in one or more of the `bad' grammars that is not also in one or more of the `good' grammars may be destroyed.

It seems likely that, in a process that may be seen as a reversal of the way in which pattern assemblies and their interconnections are created, the destruction of a pattern assembly does not mean the physical destruction of its neurons. It seems more likely that all neural connections from the pattern assembly are broken by switching off relevant synapses (Sections \ref{sp-n_destroying_neural_connections_section} and \ref{sp-n_speed_expressiveness_section}) and that its constituent neurons are retained for later use in other pattern assemblies.

\subsection{Searching for good grammars}

It must be admitted that, apart from the remarks in forgoing subsections about the creation and destruction of pattern assemblies and their inter-connections, it is unclear how, in SP-neural, one may achieve anything equivalent to the process of searching the abstract space of possible grammars that has been implemented in the SP computer model.

One possibility is to simplify things as follows. Instead of evaluating whole grammars, as in the SP computer model, it may be possible to achieve roughly the same effect by evaluating pattern assemblies in terms of their effectiveness or otherwise for the economical encoding of New information and, periodically, to discard those pattern assemblies that do badly.

\subsection{What about Hebbian learning?}\label{sp-n_hebbian_learning_section}

Readers familiar with issues in AI or neuroscience may wonder what place, if any, there may be in SP-neural for the concept of `Hebbian' learning. This idea, proposed by Hebb \cite{hebb_1949}, is that:

\begin{quote}

    ``When an axon of cell A is near enough to excite a cell B and repeatedly or persistently takes part in firing it, some growth process or metabolic change takes place in one or both cells such that A's efficiency, as one of the cells firing B, is increased.'' (p.~62).

\end{quote}

\noindent Variants of this idea are widely used in versions of `deep learning' in artificial neural networks \cite{schmidhuber_2015} and have contributed to success with such systems.\footnote{See, for example, ``Don't despair if Google's AI beats the world's best Go player'', {\em MIT Technology Review}, \href{http://bit.ly/1p7Wzb7}{bit.ly/1p7Wzb7}, 2016-03-08; and ``Google unveils neural network with `superhuman' ability to determine the location of almost any image'', {\em MIT Technology Review}, \href{http://bit.ly/1p5qmSe}{bit.ly/1p5qmSe}, 2016-02-24.}

But in \cite[Section V-D]{sp_alternatives} I have argued that:

\begin{itemize}

  \item The gradual strengthening of neural connections which is a central feature of Hebbian learning (and deep learning) does not account for the way that people can, very effectively, learn from a single occurrence or experience (sometimes called `one-trial' learning).\footnote{It may be argued that Hebbian learning may apply in such cases because a single experience may be mentally rehearsed. But that begs the question of how the one experience is remembered between when it first occurred and the first rehearsal---and likewise later on. And, while rehearsal may be helpful in some cases, it seems that there are many things we do remember after a single experience, without rehearsal.}

  \item Hebb was aware that his theory of learning with cell assemblies would not account for one-trial learning and he proposed a `reverberatory' theory for that kind of learning \cite[p.~62]{hebb_1949}. But, as noted in \cite[Section V-D]{sp_alternatives}, Milner has pointed out \cite{milner_1996} that it is difficult to understand how this kind of mechanism could explain our ability to assimilate a previously-unseen telephone number: for each digit in the number, its pre-established cell assembly may reverberate; but this does not explain memory for the {\em sequence} of digits in the number. And it is unclear how the proposed mechanism would encode a phone number in which one or more of the digits is repeated.

  \item One-trial learning is consistent with the SP theory because the direct intake and storage of sensory information is bedrock in how the system learns (Section \ref{sp-a_early_learning_section}).

  \item The SP theory can also account for the relatively slow learning of complex skills such as how to talk or how to play tennis at a high standard---because of the complexity of the abstract space of possible solutions that needs to be searched.

\end{itemize}

Does this mean that Hebbian learning is dead? Probably not:

\begin{itemize}

  \item In some forms, the phenomena of `long-term potentiation' (LPT) in neural functioning seem to be linked to Hebbian types of learning \cite[pp.~1022-1023]{squire_etal_2013}.

  \item Gradual strengthening of neural connections may have a role to play in SP-neural because some such mechanism is needed to record, at least approximately, the frequency of occurrence of neural symbols and pattern assemblies (Sections \ref{sp-a_early_learning_section} and \ref{sp-n_neural_processing_section}).

\end{itemize}

\section{The problems of speed and expressiveness in the creation of neural structures}\label{sp-n_speed_expressiveness_section}

A general issue for any neural theory of the representation and processing of knowledge, is how to account for the speed with which we can create neural structures, bearing in mind that such structures must be sufficiently versatile to accommodate the representation and processing of a wide range of different kinds of knowledge. This issue arises mainly in the following connections:

\begin{itemize}

  \item {\em One-trial learning}. In keeping with the remarks above about one-trial learning (Section \ref{sp-n_hebbian_learning_section}), it is a familiar feature of everyday life that we can see and hear something happening---a football match, a play, a conversation, and so on---and then, immediately or some time later, give a description of the event. This implies that we can lay down relevant memories at speed.

  \item {\em The learning of complex knowledge and skills}. If we accept the view of unsupervised learning which is outlined in Sections \ref{sp-a_early_learning_section}, \ref{sp-a_later_learning_section}, and \ref{sp-n_unsupervised_learning_section}, then it seems necessary to suppose that pattern assemblies are created and destroyed during the search for one or two grammars that provide a `good' description of the knowledge or skills that is being learned---and it seems likely that the creation and destruction of pattern assemblies would be fast.

  \item {\em The interpretation of sensory data}. In processes like the parsing of natural language or, more generally, understanding natural language, and in processes like pattern recognition, reasoning, and more, it seems necessary to create intermediate structures like those shown in Figure \ref{fortune_brave_multiple_alignment_figure}, and for those structures to be created at speed.

  \item {\em Speech and action}. In a similar way, it seems necessary for us to create mental structures fast in any kind of activity that requires thought, such as speaking in a way that is meaningful and comprehensible, most kinds of sport, most kinds of games, and so on.

  \item {\em Imagination}. Most people have little difficulty in imagining things they are unlikely ever to have seen---such as a cat with a coat made of grass instead of fur, or a cow with two tails. We can create such ideas fast and, if we like them well enough, we may remember them for many years.

\end{itemize}

One possible solution, which is radically different from SP-neural, is to suppose that our knowledge is stored in some chemical form such as DNA, and that the kinds of mental processes mentioned above might be mediated via the creation and modification of such chemicals. Another possibility is that learning is mediated by epigenetic mechanisms, as outlined in \cite[Section 7.4]{baars_gage_2010}. Without wishing to prejudge what the primary mechanism of learning may be, or whether perhaps there are several such mechanisms, this paper focusses on SP-neural and how it may combine speed with expressiveness, as seems to be required for the kinds of functions outlined above.

At first sight, the problem of speed in the creation of pattern assemblies and their interconnections is solved via the long-established idea that we can remember things for a few seconds via a `short-term memory'\footnote{``Short-term memory'', {\em Wikipedia}, \href{http://bit.ly/1RzAVHN}{bit.ly/1RzAVHN}, retrieved 2016-04-04.} that is distinct from `long-term memory'\footnote{``Long-term memory'', {\em Wikipedia}, \href{http://bit.ly/1M9uPhh}{bit.ly/1M9uPhh}, retrieved 2016-04-04.} and `working memory'.\footnote{``Working memory'', {\em Wikipedia}, \href{http://bit.ly/1PQq0UA}{bit.ly/1PQq0UA}, retrieved 2016-04-04.} But there is some uncertainty about the extent to which these three kinds of memory may be distinguished, one from another, and there is considerable uncertainty about how they might work, and how information may be transferred from one kind of memory to another.

As a proffered contribution to discussions in this area, the suggestion here is that, in any or all of short-term memory, working memory, and long-term memory, SP-neural may achieve the necessary speed in the creation of new structures, combined with versatility in the representation and processing of diverse kinds of knowledge, by the creation of pattern assemblies and their interconnections via the switching on and off of synapses in pre-established neural structures and their inter-connections---somewhat like the making and breaking of connections in a telephone exchange or the creation of electronic circuits in ULAs and FPGAs, as outlined in Section \ref{sp-n_creating_neural_connections_section}.

With regard to possible mechanisms for the switching on and off of synapses:

\begin{itemize}

    \item It appears that, in the entorhinal cortex between the hippocampus and the neocortex, there are neurons that can be switched ``on'' and ``off'' in an all-or-nothing manner \cite{tahvildari_etal_2007}, and we may suppose that synapses have a role to play in this behaviour.

    \item ``The efficacy of a synapse can be potentiated through at least six mechanisms'' \cite[Caption to Figure 47.10]{squire_etal_2013} and it is possible that at least one them has the necessary speed, especially since ``[Long-term potentiation] is defined as a persistent increase in synaptic strength ...~that can be induced {\em rapidly} by a brief burst of spike activity in the presynaptic afferents.'' (emphasis added) \cite[p.~1016]{squire_etal_2013}.

    \item ``[Long-term depression] is believed by many to be ...~a process whereby [Long-term potentiation] could be reversed in the hippocampus and neocortex ....'' \cite[p.~1023]{squire_etal_2013}.

    \item ``...~it is now evident that [Long-term potentiation], at least in the dentate gyrus, can either be ...~stable, lasting months or longer.'' \cite[Abstract]{abraham_2003}, although there appears to be little or no evidence with a bearing on whether or not there might be an upper limit to the duration of long-term potentiation.

    \item There is evidence that the protein kinase M$\zeta$ (PKM$\zeta$) may provide a means of turning synapses on and off, and thus perhaps storing long-term memories \cite{ogasawara_kawato_2010}.

\end{itemize}

With all these possible mechanisms, key questions are: do they act fast enough to account for the speed of the phenomena described above; and can they provide the basis for memories that can last for 50 years or more.

\section{Errors of omission, commission, and substitution}

A prominent feature of human perception is that we have a robust ability to recognise things despite disturbances of various kinds. We can, for example, recognise a car when it is partially obscured by the leaves and branches of a tree, or by falling snow or rain.

One of the strengths of SP-abstract and its realisation in the SP computer model is that, in a similar way, recognition of a New pattern or patterns is not unduly disturbed by errors of omission, commission, and substitution in those data (\cite[Chapter 6]{wolff_2006}, \cite[Section 4.2.2]{sp_extended_overview}). This is because of the way the SP computer model searches for a global optimum in the building of multiple alignments, so that it does not depend on the presence or absence of any particular feature or combination of features in the New information that is being analysed.

In its overall structure, SP-neural seems to lend itself to that kind of robustness in recognition in the face of errors in data. But the devil is in the detail. In further development of the theory, and in the development of a computer model of SP-neural, it will be necessary to clarify the details of how that kind of robustness may be achieved. In shaping this aspect of SP-neural, the principles that have been developed in SP-abstract are likely to prove useful and, with empirical evidence from brains and nervous systems, they may serve as a touchstone of success.

\section{Conclusion}

As was mentioned in the Introduction, SP-neural is a tentative and partial theory. That said, the close relationship between SP-neural and SP-abstract, the incorporation into SP-abstract of many insights from research on human perception and cognition, strengths of SP-abstract in terms of simplicity and power (Section \ref{sp-a_evaluation_theory_section}), and advantages of SP-abstract compared with other AI-related systems (Section \ref{sp-a_distinctive_features_advantages_section})---lend support to SP-neural as it is now as a conceptual model of the representation and processing of knowledge in the brain, and a promising basis for further research.

Naturally, we may have more confidence in some parts of the theory than others. Arguably, the parts that inspire most confidence are these:

\begin{itemize}

    \item {\em Neural symbols and pattern assemblies}. {\em All} knowledge is represented in the cerebral cortex with {\em pattern assemblies}, the neural equivalent of patterns in SP-abstract. Each such pattern assembly is an array of {\em neural symbols}, each of which is a single neuron or a small cluster of neurons---the neural equivalent of a symbol in SP-abstract. Topologically, each array has one or two dimensions, perhaps parallel to the surface of the cortex.

    \item {\em Information compression via the matching and unification of patterns}. As in SP-abstract, SP-neural is governed by the overarching principle that many aspects of perception and cognition may be understood in terms of information compression via the matching and unification of patterns.

    \item {\em Information compression via multiple alignment}. More specifically, SP-neural is governed by the overarching principle that many aspects of perception and cognition may be understood via a neural equivalent of the powerful concept of {\em multiple alignment}.

    \item {\em Unsupervised learning}. As in SP-abstract, unsupervised learning in SP-neural is the foundation for other kinds of learning---supervised learning, reinforcement learning, learning by imitation, learning by being told, and so on. And as in SP-abstract, unsupervised learning in SP-neural is achieved via a search through alternative grammars to find one or two that score best in terms of the compression of sensory information. As noted in Section \ref{sp-n_hebbian_learning_section}, this is quite different from the kinds of `Hebbian' learning that are popular in artificial neural networks.

    \item {\em Problems of speed and expressiveness in the creation of pattern assemblies and their interconnections}. To account for the speed with which we can assimilate new information, and the speed of other mental processes (Section \ref{sp-n_speed_expressiveness_section}), it seems necessary to suppose that pattern assemblies and their interconnections may be created from pre-existing neural structures by the making and breaking of synaptic connections, somewhat like the making and breaking of connections in a telephone exchange, or the creation of a bespoke electronic system from an `uncommitted logic array' (ULA) or a `field-programmable gate array' (FPGA).

\end{itemize}

As with SP-abstract, areas of uncertainty in SP-neural may be clarified by casting the theory in the form of a computer model and testing it to see whether or not it works as anticipated. It is envisaged that this would be part of a proposed facility for the development of the SP machine \cite{sp_proposal}, a means for researchers everywhere to explore what can be done with the SP machine and to create new versions of it.

At all stages in its development, the theory may suggest possible investigations of the workings of brains and nervous systems. And any neurophysiological evidence may have a bearing on the perceived validity of the theory and whether or how it may need to be modified.

\appendix

\section{Cell assemblies and pattern assemblies}\label{cell_pattern_assemblies_appendix}

The main differences between Hebb's \cite{hebb_1949} concept of a `cell assembly' and the SP-neural concept of a `pattern assembly' are:

\begin{itemize}

  \item The concept of a pattern assembly has had the benefit of computer modelling of SP-abstract---reducing vagueness in the theory and testing whether or not proposed mechanisms actually work as anticipated. These things would have been difficult or impossible for Hebb to do in 1949.

  \item Cell assemblies were seen largely as a vehicle for recognition, whereas, as neural realisations of SP `patterns', pattern assemblies should be able to mediate several aspects of intelligence, including recognition.

  \item Anatomically, pattern assemblies are seen as largely flat groupings of neurons in the cerebral cortex (Section \ref{sp-n_pattern_assemblies_section}), whereas cell assemblies are seen as structures in three dimensions.

  \item As described below, a fourth difference between cell assemblies and pattern assemblies is in how structures may be shared.

\end{itemize}

\begin{figure}[!htbp]
\centering
\includegraphics[width=0.6\textwidth]{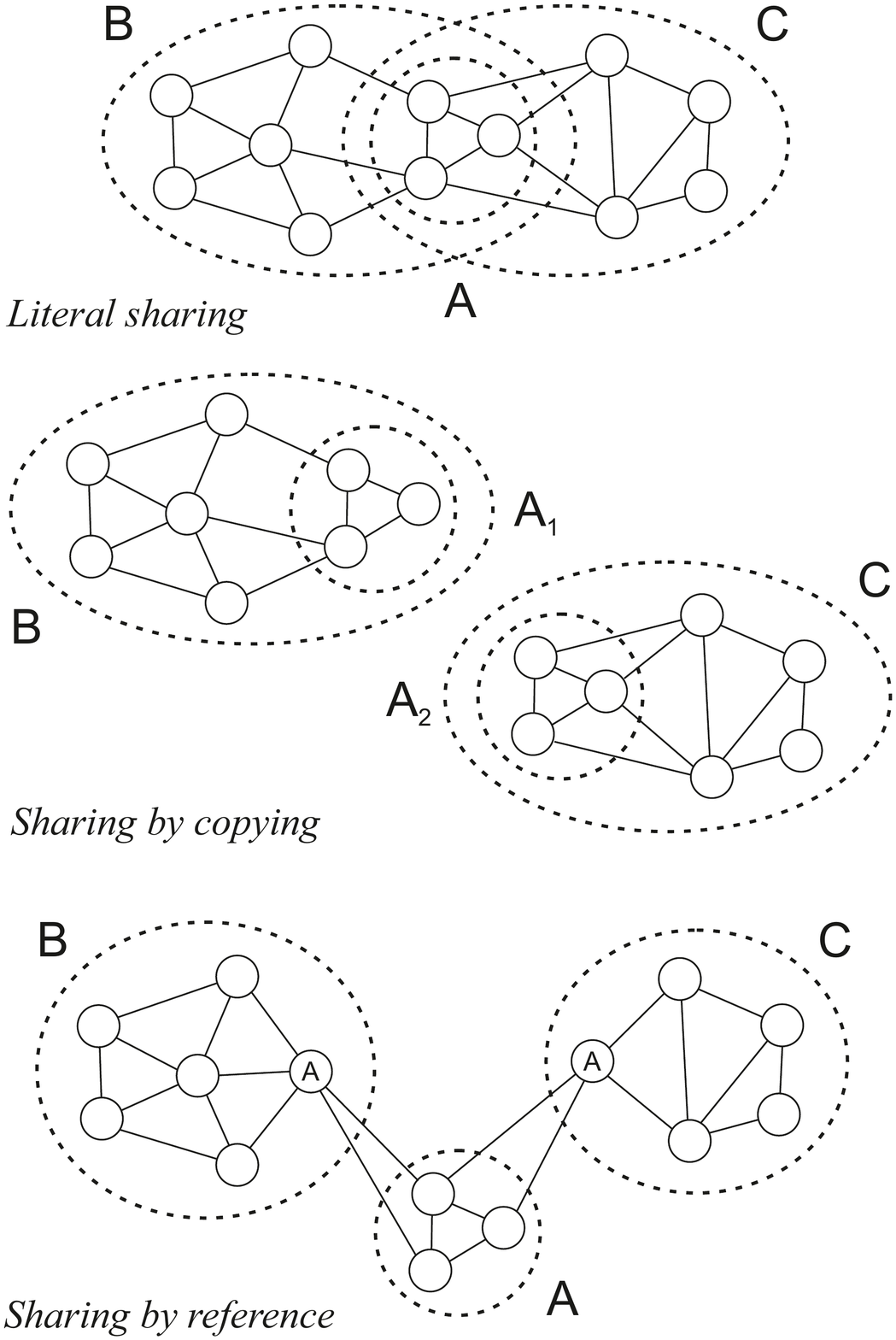}
\caption{Three possible ways in which cell assemblies may be shared, as described in the text. Adapted with permission from \cite[Figure 11.3]{wolff_2006}.}
\label{sharing_figure}
\end{figure}

With regard to the last point, possible models for sharing of structures are illustrated in Figure \ref{sharing_figure}.

In literal sharing, structures B and C in the figure both contain structure A. In sharing by copying, structures B and C each contains a copy of structure A. While in sharing by reference, structures B and C each contains a reference to structure A, in much the same way that a paper like this one contains references to other publications.

From Hebb's \cite{hebb_1949} descriptions of the cell assembly concept, it is difficult to tell which of these three possibilities are intended.

By contrast, with the concept of a pattern assembly in the SP-neural, sharing is almost always achieved by means of neural `references' between structures. For example, a noun like `table' is likely to have neural connections to the many grammatical contexts in which it may occur, as suggested by the two broken-line connections from each of `N' and `\#N' in the pattern assembly for `table' shown in Figure \ref{the_brave_neural_figure}. Notice that, in this example, the putative direction of travel of nerve impulses is not relevant---it is the neural connection that counts.

In the SP system, it is intended that literal sharing should be impossible and that sharing by copying may only occur on the relatively rare occasions when the system has failed to detect the corresponding redundancy, and not always then.

\bibliographystyle{plain}

\begin{thebibliography}{10}

\bibitem{abraham_2003}
W.~C. Abraham.
\newblock How long will long-term potentiation last?
\newblock {\em Philosophical Transactions of the Royal Society B},
  358(1432):735--744, 2003.

\bibitem{attneave_1954}
F.~Attneave.
\newblock Some informational aspects of visual perception.
\newblock {\em Psychological Review}, 61:183--193, 1954.

\bibitem{baars_gage_2010}
b.~j. baars and N.~M. Gage.
\newblock {\em Cognition, brain, and consciousness: introduction to cognitive
  neuroscience}.
\newblock Elsevier, Amserdam, second edition, 20.

\bibitem{barghout-stein_1999}
L.~Barghout-Stein.
\newblock On differences between peripheral and foveal pattern masking.
\newblock Technical report, University of California, Berkeley, 1999.
\newblock Master's thesis. \href{http://bit.ly/1SCoUO4}{bit.ly/1SCoUO4}.

\bibitem{barlow_1959}
H.~B. Barlow.
\newblock Sensory mechanisms, the reduction of redundancy, and intelligence.
\newblock In {HMSO}, editor, {\em The Mechanisation of Thought Processes},
  pages 535--559. Her Majesty's Stationery Office, London, 1959.

\bibitem{barlow_1969}
H.~B. Barlow.
\newblock Trigger features, adaptation and economy of impulses.
\newblock In K.~N. Leibovic, editor, {\em Information Processes in the Nervous
  System}, pages 209--230. Springer, New York, 1969.

\bibitem{barlow_1982}
H.~B. Barlow.
\newblock David {H}ubel and {T}orsten {W}iesel: their contribution towards
  understanding the primary visual cortex.
\newblock {\em Trends in Neuroscience}, 5(5):145--152, 1982.

\bibitem{barrow_1992}
J.~D. Barrow.
\newblock {\em Pi in the Sky}.
\newblock Penguin Books, Harmondsworth, 1992.

\bibitem{drew_atal_2013}
T.~Drew, M.~L-H. V{\"o}, and J.~M. Wolfe.
\newblock The invisible gorilla strikes again: sustained inattentional
  blindness in expert observers.
\newblock {\em Psychological Science}, 24(9):1848--1853, 2013.

\bibitem{gold_1967}
M.~Gold.
\newblock Language identification in the limit.
\newblock {\em Information and Control}, 10:447--474, 1967.

\bibitem{gross_2002}
C.~G. Gross.
\newblock Genealogy of the ``{G}randmother {C}ell''.
\newblock {\em Neuroscientist}, 8(5):512--518, 2002.

\bibitem{hebb_1949}
D.~O. Hebb.
\newblock {\em The Organization of Behaviour}.
\newblock John Wiley \& Sons, New York, 1949.

\bibitem{herculano-houzel_2012}
S.~Herculano-Houzel.
\newblock The remarkable, yet not extraordinary, human brain as a scaled-up
  primate brain and its associated cost.
\newblock {\em Proceedings of the National Academy of Sciences of the United
  States of America}, 109(Supplement 1):10661--10668, 2012.

\bibitem{huth_etal_2016}
A.~G. Huth, W.~A. de~Heer, T.~L. Griffiths, F.~E. Theunissen, and J.~L.
  Gallant.
\newblock Natural speech reveals the semantic maps that tile human cerebral
  cortex.
\newblock {\em Nature}, 532:453--458, 2016.

\bibitem{isaacson_scanziani_2011}
J.~S. Isaacson and M.~Scanziani.
\newblock How inhibition shapes cortical activity.
\newblock {\em Neuron}, 72(2):231--243, 0000.

\bibitem{mccorduck_2004}
P.~McCorduck.
\newblock {\em Machines who think: a personal inquiry into the history and
  prospects of artificial intelligence}.
\newblock A.~K.~Peters Ltd, Natick, MA, second edition, 2004.
\newblock {ISBN}: 1-56881-205-1.

\bibitem{milner_1996}
P.~M. Milner.
\newblock Neural representations: some old problems revisited.
\newblock {\em Journal of Cognitive Neuroscience}, 8(1):69--77, 1996.

\bibitem{newell_1973}
A.~Newell.
\newblock You can't play 20 questions with nature and win: projective comments
  on the papers in this symposium.
\newblock In W.~G. Chase, editor, {\em Visual Information Processing}, pages
  283--308. Academic Press, New York, 1973.

\bibitem{newell_1990}
A.~Newell, editor.
\newblock {\em Unified Theories of Cognition}.
\newblock Harvard University Press, Cambridge, Mass., 1990.

\bibitem{ogasawara_kawato_2010}
H.~Ogasawara and M.~Kawato.
\newblock The protein kinase m$\zeta$ network as a bistable switch to store
  neuronal memory.
\newblock {\em BMC Systems Biology}, 4:181--191, 2010.

\bibitem{ratliff_etal_1963}
F.~Ratliff, H.~K. Hartline, and W.~H. Miller.
\newblock Spatial and temporal aspects of retinal inhibitory interaction.
\newblock {\em Journal of the Optical Society of America}, 53:110--120, 1963.

\bibitem{schmidhuber_2015}
J.~Schmidhuber.
\newblock Deep learning in neural networks: an overview.
\newblock {\em Neural Networks}, 61:85--117, 2015.

\bibitem{shamma_1985}
S.~A. Shamma.
\newblock Speech processing in the auditory system {II}: Lateral inhibition and
  the central processing of speech evoked activity in the auditory nerve.
\newblock {\em Journal of the Acoustical Society of America}, 76(5):1622--1632,
  1985.

\bibitem{simons_ambinder_2005}
D.~J. Simons and M.~S. Ambinder.
\newblock Change blindness: theory and consequences.
\newblock {\em Current Directions in Psychological Science}, 14(1):44--48,
  2005.

\bibitem{squire_etal_2013}
L.~R. Squire, D.~Berg, F.~E. Bloom, S.~du~Lac, A.~Ghosh, and N.~C. Spitzer,
  editors.
\newblock {\em Fundamental neuroscience}.
\newblock Elsevier, Amsterdam, fourth edition, 2013.

\bibitem{stratton_1897}
G.~M. Stratton.
\newblock Upright vision and the retinal image.
\newblock {\em Psychological Review}, 4:182--187, 1897.

\bibitem{tahvildari_etal_2007}
B.~Tahvildari, E.~Frans{\'e}n, A.~A. Alonso, and M.~E. Hasselmo.
\newblock Switching between ``on'' and ``off'' states of persistent activity in
  lateral entorhinal layer {III} neurons.
\newblock {\em Hippocampus}, 17:257--263, 2007.

\bibitem{von_bekesy_1967}
G.~{von B{\'e}k{\'e}sy}.
\newblock {\em Sensory Inhibition}.
\newblock Princeton University Press, Princeton, NJ, 1967.

\bibitem{wolff_1988}
J.~G. Wolff.
\newblock Learning syntax and meanings through optimization and distributional
  analysis.
\newblock In Y.~Levy, I.~M. Schlesinger, and M.~D.~S. Braine, editors, {\em
  Categories and Processes in Language Acquisition}, pages 179--215. Lawrence
  Erlbaum, Hillsdale, NJ, 1988.
\newblock \href{http://bit.ly/ZIGjyc}{bit.ly/ZIGjyc}.

\bibitem{wolff_medical_diagnosis}
J.~G. Wolff.
\newblock Medical diagnosis as pattern recognition in a framework of
  information compression by multiple alignment, unification and search.
\newblock {\em Decision Support Systems}, 42:608--625, 2006.
\newblock \href{http://bit.ly/XE7pRG}{bit.ly/XE7pRG}, arXiv:1409.8053 [cs.AI].

\bibitem{wolff_2006}
J.~G. Wolff.
\newblock {\em Unifying Computing and Cognition: the {SP} Theory and Its
  Applications}.
\newblock CognitionResearch.org, Menai Bridge, 2006.
\newblock {ISBN}s: 0-9550726-0-3 (ebook edition), 0-9550726-1-1 (print
  edition). Distributors, including Amazon.com, are detailed on
  \href{http://bit.ly/WmB1rs}{bit.ly/WmB1rs}.

\bibitem{wolff_sp_intelligent_database}
J.~G. Wolff.
\newblock Towards an intelligent database system founded on the {SP} theory of
  computing and cognition.
\newblock {\em Data \& Knowledge Engineering}, 60:596--624, 2007.
\newblock \href{http://bit.ly/Yg2onp}{bit.ly/Yg2onp}, arXiv:cs/0311031 [cs.DB].

\bibitem{sp_extended_overview}
J.~G. Wolff.
\newblock The {SP} theory of intelligence: an overview.
\newblock {\em Information}, 4(3):283--341, 2013.
\newblock \href{http://bit.ly/1hz0lFE}{bit.ly/1hz0lFE}.

\bibitem{sp_vision}
J.~G. Wolff.
\newblock Application of the {SP} theory of intelligence to the understanding
  of natural vision and the development of computer vision.
\newblock {\em SpringerPlus}, 3(1):552--570, 2014.
\newblock \href{http://bit.ly/1scmpV9}{bit.ly/1scmpV9}.

\bibitem{sp_autonomous_robots}
J.~G. Wolff.
\newblock Autonomous robots and the {SP} theory of intelligence.
\newblock {\em IEEE Access}, 2:1629--1651, 2014.
\newblock \href{http://bit.ly/1zrSemu}{bit.ly/1zrSemu}.

\bibitem{sp_big_data}
J.~G. Wolff.
\newblock Big data and the {SP} theory of intelligence.
\newblock {\em IEEE Access}, 2:301--315, 2014.
\newblock \href{http://bit.ly/1jGWXDH}{bit.ly/1jGWXDH}. This article, with
  minor revisions, is to be reproduced in Fei Hu (Ed.), {\em Big Data: Storage,
  Sharing, and Security (3S)}, Taylor \& Francis LLC, CRC Press, 2016, pp.
  143--170.

\bibitem{sp_foundations}
J.~G. Wolff.
\newblock Information compression, intelligence, computing, and mathematics.
\newblock Technical report, CognitionResearch.org, 2014.
\newblock \href{http://bit.ly/1jEoECH}{bit.ly/1jEoECH}, arXiv:1310.8599
  [cs.AI]. Submitted for publication.

\bibitem{sp_benefits_apps}
J.~G. Wolff.
\newblock The {SP} theory of intelligence: benefits and applications.
\newblock {\em Information}, 5(1):1--27, 2014.
\newblock \href{http://bit.ly/1lcquWF}{bit.ly/1lcquWF}.

\bibitem{sp_alternatives}
J.~G. Wolff.
\newblock The {SP} theory of intelligence: its distinctive features and
  advantages.
\newblock {\em IEEE Access}, 4:216--246, 2016.
\newblock \href{http://bit.ly/21gv2jT}{bit.ly/21gv2jT}.

\bibitem{sp_proposal}
J.~G. Wolff and V.~Palade.
\newblock Proposal for the creation of a research facility for the development
  of the {SP} machine.
\newblock Technical report, CognitionResearch.org, 2015.
\newblock \href{http://bit.ly/1zZjjIs}{bit.ly/1zZjjIs}, arXiv:1508.04570
  [cs.AI].

\end{thebibliography}

\end{document}